\pdfoutput=1
\documentclass[10pt]{article}

\PassOptionsToPackage{numbers,compress,sort}{natbib}
\usepackage{natbib}
\usepackage[utf8]{inputenc}
\usepackage[T1]{fontenc}
\usepackage{lmodern}
\usepackage[a4paper,margin=1in]{geometry}
\usepackage{url}
\usepackage{booktabs}
\usepackage{amsmath,amssymb,amsfonts}
\usepackage{nicefrac}
\usepackage{xspace}
\usepackage[protrusion=true,expansion=true,final]{microtype}
\usepackage{graphicx}
\usepackage{tabularx}
\usepackage{adjustbox}
\usepackage{float}
\usepackage[section]{placeins}

\usepackage[table]{xcolor}
\usepackage{algorithm}
\usepackage{algpseudocode}
\usepackage{caption}
\usepackage[colorlinks=true,linkcolor=black,citecolor=black,urlcolor=blue]{hyperref}
\usepackage{doi}
\usepackage[most]{tcolorbox}

\definecolor{pamirCard}{RGB}{247,248,250}   
\definecolor{pamirInk}{HTML}{12303A}        
\definecolor{pamirFocal}{HTML}{0B7A83}      

\tcbset{
  colback=pamirCard,
  colframe=black!15,
  boxrule=0.5pt,
  arc=6pt,
  left=16pt,right=16pt,top=14pt,bottom=14pt,
  enhanced,
}

\newcommand{\field}[2]{%
  \par\noindent\makebox[5.4em][l]{\small\color{black!55}\textsc{#1}}#2\par\addvspace{2pt}}
\newcommand{\stat}[2]{%
  {\Large\bfseries\color{pamirInk}#1\par}\vspace{1pt}%
  {\scriptsize\color{black!60}#2\par}}

\newcommand{\provisional}[1]{\textcolor{red}{#1}}
\newcommand{\provurl}[1]{\href{#1}{\textcolor{red}{\nolinkurl{#1}}}}

\newcommand{\pamir}{PaMIR}
\newcommand{\code}[1]{\texttt{#1}}

\hypersetup{
pdftitle={PaMIR: The Largest Open Benchmark of Public Credit-Default Datasets},
pdfsubject={cs.LG, q-fin.RM},
pdfauthor={Mikhail Liashkov, Ilyas Varshavskiy, Bonu Boboeva, Shuhrat Khalilbekov, Azizjon Azimi},
pdfkeywords={credit risk, default prediction, benchmark, living benchmark, tabular data, label delay, data leakage, synthetic data},
}

\begin{document}

\vspace*{-1.1cm}
\begin{tcolorbox}

\includegraphics[height=0.85cm]{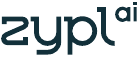}

\vspace{0.7em}
{\LARGE\bfseries PaMIR: Open Benchmark of\\[2pt] Public Credit-Default Datasets\par}

\vspace{0.7em}
{\normalsize
Mikhail Liashkov\,\href{https://orcid.org/0009-0002-3734-8441}{\includegraphics[height=0.75em]{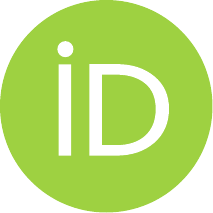}}\textsuperscript{1},
Ilyas Varshavskiy\,\href{https://orcid.org/0000-0003-4924-2097}{\includegraphics[height=0.75em]{orcid.pdf}}\textsuperscript{1},
Bonu Boboeva\textsuperscript{1},
Shuhrat Khalilbekov\textsuperscript{1,*},
Azizjon~Azimi\textsuperscript{1}\par}

\vspace{0.3em}
{\small\color{black!65}\textsuperscript{1}zypl.ai\quad \textsuperscript{*}Corresponding author: \href{mailto:shuhrat.khalilbekov@zypl.ai}{shuhrat.khalilbekov@zypl.ai}\par}

\vspace{1.1em}
We release \textbf{\pamir{}} (Public Arrival-ordered Measurement for Inference in Risk), an open benchmark for \textbf{credit-default prediction when labels are scarce and arrive late}. The field's reference benchmark studies use eight datasets each, only two or four of them public. \pamir{} brings together \textbf{19 public datasets} with binary default labels --- 1.24M loans, firms and card accounts from nine countries --- \textbf{rebuilt from pinned source snapshots} by one leakage-audited recipe and never redistributed; to our knowledge it is the one of its kind as of today. Every model is a single function, scored under a \textbf{repeated i.i.d.\ split} and a \textbf{label-delayed stream} in which each application is scored on arrival, with AUC reported by label budget; fleet means are withheld unless every dataset is scored. A \textbf{synthetic-data harness} tests generated training rows without letting a generator see held-out rows. This report describes release 0.4.0 of this \textbf{living benchmark}.

\vspace{0.9em}
\field{Release}{0.4.0 --- draft of \today}
\field{Install}{\provisional{\texttt{pip install pamir}}}
\field{Code}{\provurl{https://github.com/zypl-ai/pamir}}
\field{Docs}{\provurl{https://pamir-docs.pages.dev}}
\field{HF card}{\provisional{Hugging Face dataset card --- URL to be announced}}
\field{License}{Code: Apache-2.0; each dataset stays under its source's licence (Section~\ref{sec:licensing})}

\end{tcolorbox}

\vspace{0.35cm}
{\small\color{black!55}\textsc{PaMIR 0.4 at a glance}}\par\vspace{2pt}
{\setlength{\tabcolsep}{3pt}%
\noindent\begin{tabularx}{\linewidth}{@{}*{7}{>{\centering\arraybackslash}X}@{}}
\stat{19}{public\newline datasets} &
\stat{1.24M}{rows\newline in total} &
\stat{9}{named\newline countries} &
\stat{7}{product\newline types} &
\stat{3--41\%}{default\newline rates} &
\stat{7}{candidates\newline rejected} &
\stat{0}{tables\newline redistributed} \\
\end{tabularx}}

\vspace{0.1cm}
\begin{figure}[H]
\centering
\includegraphics[width=\linewidth]{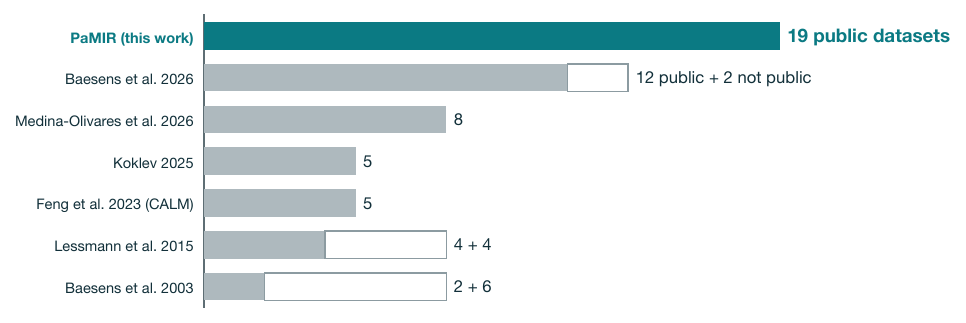}
\caption{\textbf{\pamir{} pools more public credit-default datasets than any prior multi-dataset study.} Each bar counts the datasets with a binary default, bankruptcy or financial-distress target that one study uses together; the filled part is public, the hollow part proprietary or restricted. The counting convention and the sources are given in Section~\ref{sec:largest} and Table~\ref{tab:credit}.}
\label{fig:prior}
\end{figure}

\newpage
{\setcounter{tocdepth}{2}\tableofcontents}
\newpage

\section{Introduction}
\label{sec:intro}

Credit scoring is one of the oldest applications of statistical classification \citep{hand1997, thomas2017}, and it remains one of the settings in which small differences in discrimination translate directly into money. Methodological comparisons in the field, however, rest on a narrow empirical base. Systematic reviews of the credit-scoring literature find that it is dominated by a few small UCI tables such as the German and Australian credit data \citep{louzada2016, markov2022}, and that papers published between 2016 and 2020 used 2.75 datasets on average \citep{markov2022}. The best-known benchmark studies used eight datasets each, of which only two and four respectively are public \citep{baesens2003, lessmann2015}, and the largest recent multi-dataset study uses fourteen, twelve of them public \citep{baesens2026foundation}.

Assembling a larger collection is harder than it appears, for four reasons. First, \emph{provenance is scattered}: the same underlying loan book circulates as several re-uploads with different columns, row counts and declared licences. Second, \emph{preprocessing is rebuilt per study}, so two papers that name the same dataset rarely evaluate on the same table. Third, many public credit tables contain \emph{post-outcome columns} --- recovery amounts, charged-off principal, days delinquent --- that are recorded after the outcome they are meant to predict; models trained on them report inflated performance that no lender could obtain at decision time \citep{kaufman2012, kapoor2023}. Audits of general tabular benchmarks have found such leaks in widely used datasets \citep{rubachev2025tabred}. Fourth, general-purpose tabular benchmarks \citep{bischl2021openml, grinsztajn2022, erickson2025tabarena, lee2026multitab, purucker2026beyondarena} contain few credit datasets and evaluate on random, grouped or time-based splits, whereas a deployed credit model is trained only on outcomes that have matured --- months or years after origination --- and starts with no labels at all. A time-based split orders the data by calendar, but unless it leaves a gap as long as the maturation period, its training set contains outcomes that would not yet have been observed when the test period began (Section~\ref{sec:otherwork}).

\pamir{} is built around this setting: credit-default prediction when labels are scarce and arrive late. It addresses the first three problems directly and the fourth through its streaming protocol, in which every application is scored when it arrives by a model fitted on the outcomes that have matured by then, so that the delay binds for every row and performance can be read against the number of labels a model has seen. With 19 public credit-default datasets, \pamir{} is also, to our knowledge, the largest open collection of its kind: more than twice the eight datasets of each reference benchmark study, only two and four of which are public, and more than the twelve public datasets of the largest multi-dataset study we found. Section~\ref{sec:largest} states how we count, where we looked and how robust the claim is. Because a model enters \pamir{} as one Python function, the same 19 tables can serve as a credit track for the tabular foundation models now compared on general benchmarks, in which credit appears as a handful of tasks --- BeyondArena, for example, has eleven credit-default tasks among 142 \citep{purucker2026beyondarena} --- and for the question of \citet{baesens2026foundation}, whether such models change credit-risk prediction. \pamir{} is a \emph{living} benchmark: new datasets enter through versioned releases as audited recipes (Section~\ref{sec:living}).\footnote{The acronym reads \emph{Public Arrival-ordered Measurement for Inference in Risk}. It also names the Pamir mountains of Central Asia, the ``Roof of the World'' --- a place to stress-test credit models.} Its contributions are:
\begin{enumerate}
  \item \textbf{An open, reproducible collection of 19 public credit-default datasets} (Table~\ref{tab:datasets}, Figures~\ref{fig:prior} and~\ref{fig:landscape}) covering consumer, peer-to-peer, credit-card, vehicle, small-business and corporate lending, rebuilt from pinned source snapshots by a leakage-audited recipe: a semantic day-zero cut, eight leakage guards each tested against a reconstruction of the leak it was written for, and a frozen contract with SHA-256 digests that every download must reproduce (Section~\ref{sec:collection}). The collection is positioned against prior multi-dataset studies in Table~\ref{tab:credit}; the data are fetched from their origin and never redistributed.
  \item \textbf{Protocols for scarce and delayed labels behind one model interface}: a label-delayed stream in which each application is scored on arrival, with AUC reported by label budget, and a repeated i.i.d.\ split for comparison with general benchmarks; together with \textbf{fleet summaries that refuse to average} over a partial run, over a table that failed its contract, or over scores that depend on the other rows they were computed with (Section~\ref{sec:protocols}).
  \item \textbf{A leakage-controlled harness for synthetic-data augmentation}, the remedy most often proposed for scarce labels: generators are fitted inside each fold, utility is reported as a paired delta, fidelity by a three-layer battery, and every fold is probed for reproduced held-out rows (Section~\ref{sec:synthetic}).
\end{enumerate}

All three are released as one open-source package under Apache-2.0, with versioned releases, a recipe-based path for adding datasets, a command-line interface and machine-readable documentation for language-model agents (Sections~\ref{sec:living} and~\ref{sec:software}).

This report describes release 0.4.0. Compared with release 0.3.0 it pins every source snapshot and makes the data contract binding, scores streaming applications on arrival by default (the 0.3.0 rule remains available as \code{mode="refresh"}), reports AUC by label budget, and gives reference results for two baselines (Section~\ref{sec:baselines}). Section~\ref{sec:limitations} lists what the current version does not measure, and the authors' conflict of interest is stated before the references.

\section{Related work}
\label{sec:related}

\subsection{Multi-dataset credit studies, and why \pamir{} is the largest}
\label{sec:largest}

\citet{baesens2003} and \citet{lessmann2015} established multi-dataset evaluation as the standard in credit scoring, and \citet{gunnarsson2021} extended it to deep learning. \citet{hand2006} argued that apparent progress in classifier technology is often smaller than the uncertainty introduced by population drift and by the evaluation design itself, which motivates \pamir{}'s emphasis on protocol rather than on model rankings. We call \pamir{} the largest open benchmark of public credit-default datasets; this section states what the claim means and how we checked it.

\paragraph{What we count.} A dataset counts if it has a binary default, bankruptcy or financial-distress target and is used together with the others in one study or benchmark. We count datasets rather than rows, and we count public and proprietary datasets separately, because only public datasets can be reproduced by others. Loss-given-default (regression) datasets and non-default tasks such as fraud detection or insurance claims are not counted.

\paragraph{Where we looked.} As of October 2026 we examined systematic reviews of the credit-scoring literature \citep{louzada2016, markov2022}, the reference benchmark studies \citep{baesens2003, lessmann2015}, the general tabular benchmarks discussed in Section~\ref{sec:otherwork} together with the curation records of BeyondArena \citep{datafoundry2026}, a benchmark for language models in credit and risk assessment \citep{feng2023calm}, and recent multi-dataset credit studies \citep{koklev2025, baesens2026foundation, medinaolivares2026findr}.

\paragraph{What we found.} Table~\ref{tab:credit} and Figure~\ref{fig:prior} summarize the counts. A typical credit-scoring paper uses 2.75 datasets \citep{markov2022}. The reference studies of \citet{baesens2003} and \citet{lessmann2015} use eight each, of which two and four respectively are public. The CALM benchmark for language models contains nine risk-assessment tasks, five of which are credit-scoring or financial-distress datasets, subsampled and converted into prompts \citep{feng2023calm}. Recent studies of monotone gradient boosting and of fair semi-structured regression use five and eight public datasets \citep{koklev2025, medinaolivares2026findr}. The largest study we found, a comparison of tabular foundation models with established credit-risk learners \citep{baesens2026foundation}, uses fourteen probability-of-default datasets, twelve of them public; three of those twelve (HMEQ, Thomas and Home Credit) are datasets that \pamir{}'s audit excluded (Section~\ref{sec:audit}). Among general benchmarks, BeyondArena has the most credit tasks: eleven of its 142 datasets have a binary default or bankruptcy target \citep{purucker2026beyondarena}, and three of them (HMEQ, HELOC and Home Credit) are also among \pamir{}'s exclusions. \pamir{}'s 19 public datasets exceed every one of these counts. To our knowledge, none of these studies provides a harness that rebuilds every table from its original source under a leakage audit, which is what \pamir{} is built around.

\begin{figure}[tp]
\centering
\includegraphics[width=\linewidth]{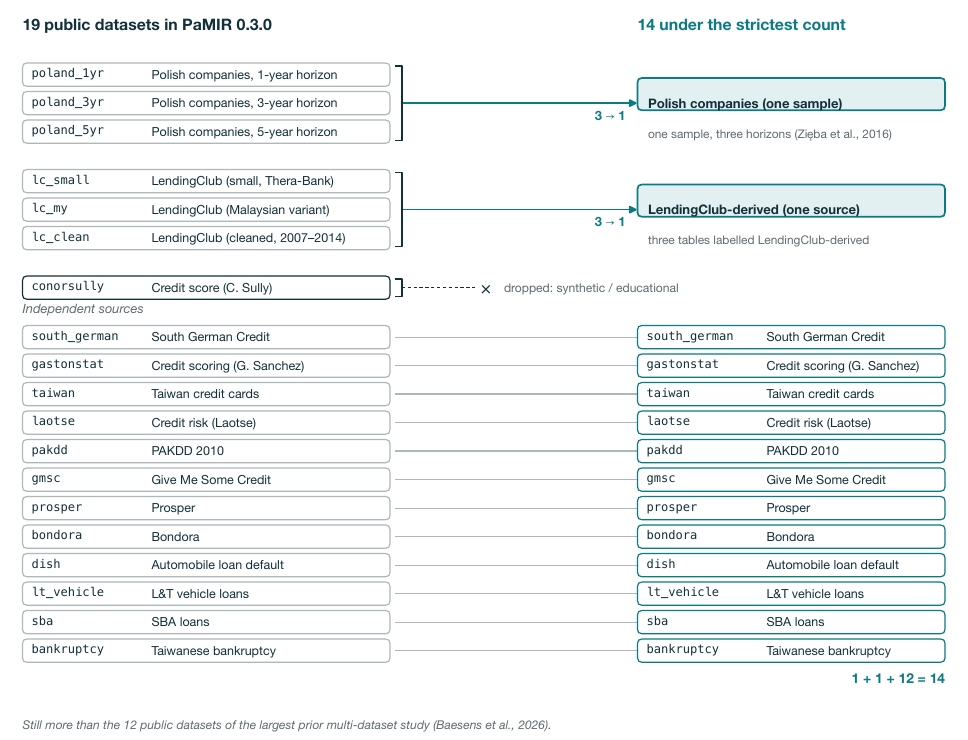}
\caption{\textbf{Under the strictest count the 19 datasets reduce to 14 independent sources, still more than the twelve public datasets of the largest prior study.} The three Polish sets are one sample observed at three forecasting horizons \citep{zieba2016} and count once; the three tables labelled as LendingClub-derived count once; the synthetic \code{conorsully} is dropped. The remaining twelve datasets are counted as they are.}
\label{fig:independent}
\end{figure}

\paragraph{How robust the claim is.} Nineteen datasets are not nineteen independent sources: the three Polish sets are one sample observed at three horizons, three tables are labelled as LendingClub-derived, and \code{conorsully} is synthetic. Under the strictest count --- merging the Polish horizons and the LendingClub tables and dropping the synthetic set --- \pamir{} retains 14 datasets (Figure~\ref{fig:independent}), still more than the twelve public datasets of the largest study above. The claim is dated: it reflects the literature as of October 2026, and we will revise it if a larger public collection is brought to our attention.

\begin{table}[t]
\centering
\caption{\textbf{No prior multi-dataset study uses as many public credit-default datasets as \pamir{}.} \emph{Default datasets} counts datasets with a binary default, bankruptcy or financial-distress target used together in one study; \emph{public} counts those that others can obtain. Status as of October 2026.}
\label{tab:credit}
\small
\begin{adjustbox}{max width=\linewidth}
\begin{tabular}{@{}lrrl@{}}
\toprule
Study & Default datasets & Public & Purpose \\
\midrule
Typical paper, 2016--2020 \citep{markov2022} & 2.75 (mean) & -- & various \\
\citet{baesens2003} & 8 & 2 & classifier comparison \\
\citet{lessmann2015} & 8 & 4 & classifier comparison \\
\citet{feng2023calm} (CALM) & 5$^{a}$ & 5 & evaluation of language models \\
\citet{koklev2025} & 5 & 5 & monotone gradient boosting \\
\citet{baesens2026foundation} & 14$^{b}$ & 12 & tabular foundation models \\
\citet{purucker2026beyondarena} (BeyondArena) & 11$^{c}$ & 11 & general tabular benchmark \\
\citet{medinaolivares2026findr} & 8 & 8 & fair semi-structured regression \\
\midrule
\rowcolor{pamirFocal!10}\textbf{\pamir{} (this work)} & \textbf{19} & \textbf{19} & open, living benchmark \\
\bottomrule
\end{tabular}
\end{adjustbox}

\vspace{2pt}
{\footnotesize $^{a}$ Plus four non-default tasks (fraud detection, insurance claims). $^{b}$ Plus seven loss-given-default datasets. $^{c}$ Of 142 datasets; HMEQ was retired from the benchmark in September 2026 \citep{datafoundry2026}.}
\end{table}

\subsection{Other benchmarks and related work}
\label{sec:otherwork}

\paragraph{General tabular benchmarks.}
OpenML-CC18 \citep{bischl2021openml} curated 72 classification datasets; \citet{grinsztajn2022} showed on a curated benchmark that tree ensembles remain ahead of deep models on medium-sized tabular data; TabArena \citep{erickson2025tabarena} is a living benchmark of 51 curated datasets with a maintained leaderboard; MultiTab \citep{lee2026multitab} organizes 196 datasets by data characteristics. All four evaluate on random splits and include only a handful of credit tasks. BeyondArena \citep{purucker2026beyondarena} extends this line to 142 datasets with i.i.d., temporal and grouped tasks, curated with the DataFoundry framework, and argues that discipline-specific evaluations remain inaccessible to model developers because their software and protocols are fragmented; the technical report of TabPFN-3.5 already evaluates on it \citep{jaeger2026tabpfn}. TabReD \citep{rubachev2025tabred} collects eight industry-grade datasets with time-based splits and shows that temporal evaluation changes method rankings; TableShift \citep{gardner2023tableshift} provides 15 binary tasks with explicit distribution shifts. \pamir{} is narrower in domain than these benchmarks and broader within it, and is meant to be run as a discipline-specific track next to them: one function, one installation, 19 credit tables. Its streaming protocol targets a constraint none of them models: labels that mature with a delay while the model must keep scoring. Table~\ref{tab:benchmarks} summarizes the comparison.

\paragraph{Time-based splits and label delay.}
Temporal tasks in general benchmarks split by calendar date. In the two temporal credit tasks of BeyondArena, LendingClub loans issued before 2016 train the model and loans issued in 2016 test it, and Home Credit applications are split on 1 May 2020; neither split leaves a gap between the training and the test period \citep{datafoundry2026}. A loan issued shortly before the cut-off resolves after it, so its outcome would not have been known when the test period began, and when only loans with a terminal status are kept, as in the LendingClub task, the most recent training cohort over-represents early defaults and early repayments. Such splits measure robustness to calendar drift, not the delay between scoring and outcome. \pamir{}'s streaming protocol has the opposite profile --- it has no calendar (Section~\ref{sec:limitations}) but makes the delay explicit --- so the two kinds of evaluation complement each other.

\begin{table}[t]
\centering
\caption{\textbf{Among these benchmarks, only \pamir{} is credit-specific and evaluates under label delay.} Counts are as reported by each benchmark's paper; credit-specific studies are compared in Table~\ref{tab:credit}.}
\label{tab:benchmarks}
\small
\begin{adjustbox}{max width=\linewidth}
\begin{tabular}{@{}lrll@{}}
\toprule
Benchmark & Datasets & Domain & Evaluation \\
\midrule
OpenML-CC18 \citep{bischl2021openml} & 72 & general & i.i.d.\ cross-validation \\
TabArena \citep{erickson2025tabarena} & 51 & general & i.i.d.\ cross-validation \\
BeyondArena \citep{purucker2026beyondarena} & 142 & general & i.i.d., temporal and grouped splits \\
MultiTab \citep{lee2026multitab} & 196 & general & i.i.d.\ splits \\
TableShift \citep{gardner2023tableshift} & 15 & general & explicit domain shifts \\
TabReD \citep{rubachev2025tabred} & 8 & industrial & time-based splits \\
\midrule
\rowcolor{pamirFocal!10}\textbf{\pamir{} (this work)} & \textbf{19} & credit default & i.i.d.\ splits and label-delayed streaming \\
\bottomrule
\end{tabular}
\end{adjustbox}
\end{table}

\paragraph{Delayed labels and stream evaluation.}
Prequential evaluation of stream learners is well studied \citep{gama2013}, as is learning when labels arrive late \citep{kuncheva2008, zliobaite2010}. \citet{grzenda2020} formalized evaluation protocols for delayed labelling, and \citet{csaba2024labeldelay} showed in online continual learning that a label delay of $d$ steps causes a degradation that extra compute does not recover. \pamir{} adopts the same structure --- features now, labels after a delay --- but applies it to credit portfolios, where the delay is the maturation period of a loan, and couples it with a refit schedule driven by resolved defaults rather than by wall time.

\paragraph{Synthetic tabular data.}
Generators for mixed-type tables range from copulas and the Synthetic Data Vault \citep{patki2016} to CTGAN \citep{xu2019ctgan} and latent or mixed-type diffusion models \citep{zhang2024tabsyn, shi2025tabdiff}. Their evaluation mixes marginal fidelity, dependence fidelity, classifier two-sample tests \citep{lopezpaz2017c2st}, privacy proxies and downstream utility, often computed on a split the generator has seen. \pamir{}'s synthetic module does not propose a generator; it fixes the protocol under which a claimed utility gain can be believed.

\section{The \pamir{} collection}
\label{sec:collection}

\subsection{Inclusion criteria}

A dataset enters \pamir{} if it meets five conditions: (i) it is a single row-per-obligor table, or a source that the recipe reduces to one by a documented join or aggregation; (ii) it has a binary default target; (iii) it is publicly downloadable without a non-disclosure agreement or institutional access; (iv) it has at least 1,000 rows; and (v) it has a default rate of at least 3\%, which leaves enough positives for a streaming evaluation to resolve.

\subsection{Composition}

\begin{figure}[t]
\centering
\includegraphics[width=\linewidth]{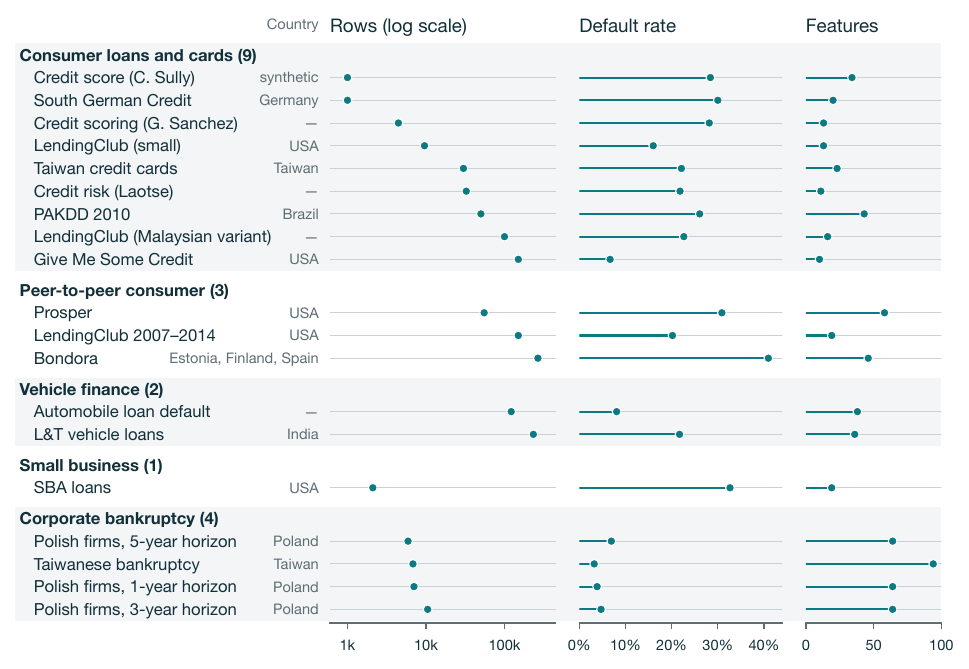}
\caption{\textbf{The 19 datasets span five product families, 1k--266k rows, default rates from 3.2\% to 40.9\% and 10 to 94 features.} One row per dataset, in the order of Table~\ref{tab:datasets}; the columns give the rows of the harmonized table (log scale), the default rate and the number of features, and the grey label gives the country of origin (a dash where the source names none). Short names are abbreviated from the catalog titles; Table~\ref{tab:datasets} gives the package identifiers.}
\label{fig:landscape}
\end{figure}

Table~\ref{tab:datasets} lists the 19 datasets and Figure~\ref{fig:landscape} shows their size, default rate and width side by side. Together they hold 1,237,550 rows and 281,766 defaults (22.8\% overall). Datasets range from 1,000 to 266,482 rows and from 10 to 94 features. Nine countries are named --- Brazil, Estonia, Finland, Germany, India, Poland, Spain, Taiwan and the USA --- and the remaining sources are labelled \emph{unspecified}, \emph{unknown} or \emph{synthetic}. The seven product types group into five families: consumer lending (including revolving credit and credit cards), peer-to-peer consumer lending, vehicle finance, small-business lending and corporate bankruptcy. The three lowest default rates (3.2\%--4.7\%) belong to corporate bankruptcy sets, and the highest (40.9\%) to a peer-to-peer book. The three Polish sets share one source and differ in forecasting horizon (one, three and five years) \citep{zieba2016}.

\begin{table}[t]
\centering
\caption{\textbf{The PaMIR collection (v0.4.0): 19 datasets, 1,237,550 rows and 281,766 defaults.} Datasets are grouped by product family. Rows, features, defaults and default rate (DR) are those of the harmonized table, validated against \texttt{expected.json}. \emph{Access} is where the recipe fetches the raw file; $^\ast$ marks the seven sources outside Kaggle, which \texttt{pamir.download\_open()} fetches. \emph{License} is as declared by the downloaded source; $^\dagger$ marks a CC0 declaration set by a Kaggle re-uploader rather than by the rights holder (Section~\ref{sec:licensing}). EE, FI, ES: Estonia, Finland, Spain.}
\label{tab:datasets}
\small
\setlength{\tabcolsep}{4pt}
\begin{adjustbox}{max width=\linewidth}
\begin{tabular}{@{}lllrrrrll@{}}
\toprule
id & product & geography & rows & feat. & defaults & DR (\%) & access & license \\
\midrule
\texttt{conorsully} & Consumer & synthetic & 1,000 & 34 & 284 & 28.4 & Kaggle & CC0 \\
\texttt{south\_german} & Consumer & Germany & 1,000 & 20 & 300 & 30.0 & UCI$^\ast$ & CC-BY-4.0 \\
\texttt{gastonstat} & Consumer & unknown & 4,454 & 13 & 1,254 & 28.1 & GitHub$^\ast$ & none stated \\
\texttt{lc\_small} & Consumer & USA & 9,578 & 13 & 1,533 & 16.0 & Kaggle & ODbL \\
\texttt{taiwan} & Credit card & Taiwan & 30,000 & 23 & 6,636 & 22.1 & Kaggle & CC0$^\dagger$ \\
\texttt{laotse} & Consumer & unspec. & 32,581 & 11 & 7,108 & 21.8 & Kaggle & CC0$^\dagger$ \\
\texttt{pakdd} & Consumer & Brazil & 50,000 & 43 & 13,041 & 26.1 & GitHub$^\ast$ & none stated \\
\texttt{lc\_my} & Consumer & unspec. & 100,000 & 16 & 22,639 & 22.6 & Kaggle & unknown \\
\texttt{gmsc} & Consumer (revolving) & USA & 150,000 & 10 & 10,026 & 6.7 & Kaggle & unknown \\
\midrule
\texttt{prosper} & P2P consumer & USA & 55,084 & 58 & 17,010 & 30.9 & Kaggle & CC0$^\dagger$ \\
\texttt{lc\_clean} & P2P consumer & USA & 150,000 & 19 & 30,273 & 20.2 & HF$^\ast$ & none stated \\
\texttt{bondora} & P2P consumer & EE, FI, ES & 266,482 & 46 & 109,124 & 40.9 & Kaggle & CC0$^\dagger$ \\
\midrule
\texttt{dish} & Vehicle finance & unspec. & 121,856 & 38 & 9,845 & 8.1 & Kaggle & CC0$^\dagger$ \\
\texttt{lt\_vehicle} & Vehicle finance & India & 233,154 & 36 & 50,611 & 21.7 & Kaggle & other \\
\midrule
\texttt{sba} & Small business & USA & 2,102 & 19 & 686 & 32.6 & Kaggle & CC0$^\dagger$ \\
\midrule
\texttt{poland\_5yr} & Corporate & Poland & 5,910 & 64 & 410 & 6.9 & UCI$^\ast$ & CC-BY-4.0 \\
\texttt{bankruptcy} & Corporate & Taiwan & 6,819 & 94 & 220 & 3.2 & Kaggle & \copyright{} authors \\
\texttt{poland\_1yr} & Corporate & Poland & 7,027 & 64 & 271 & 3.9 & UCI$^\ast$ & CC-BY-4.0 \\
\texttt{poland\_3yr} & Corporate & Poland & 10,503 & 64 & 495 & 4.7 & UCI$^\ast$ & CC-BY-4.0 \\
\midrule
\textbf{Total} & 7 product types & 9 countries & 1,237,550 & & 281,766 & 22.8 & & \\
\bottomrule
\end{tabular}
\end{adjustbox}
\end{table}

\subsection{Provenance, retrieval and validation}

\begin{figure}[tp]
\centering
\includegraphics[width=\linewidth]{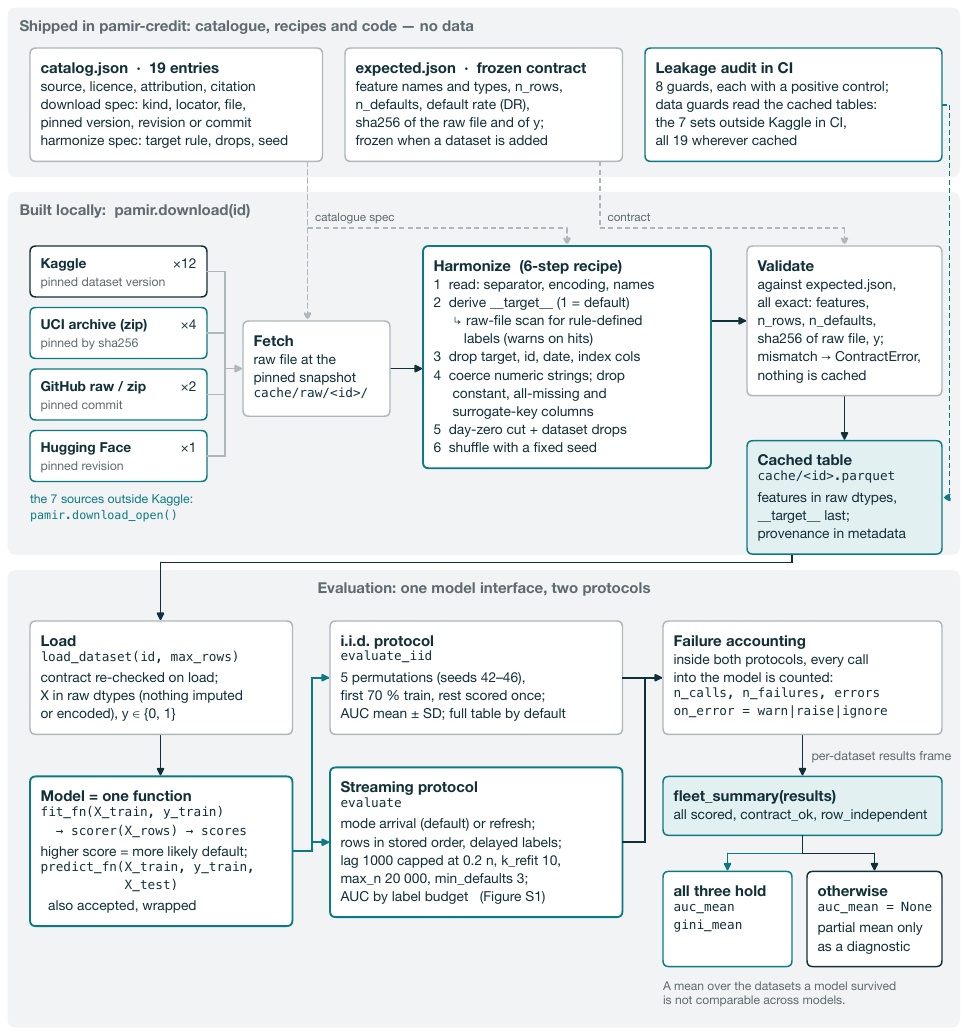}
\caption{\textbf{\pamir{} ships metadata and code, rebuilds every table locally from a pinned source snapshot, and reports a fleet mean only for a complete, valid run.} Top: what the package contains --- the catalogue with each dataset's source, licence, citation, pinned download specification and harmonization specification; the frozen \code{expected.json} contract; and the leakage guards run in continuous integration. Middle: \code{pamir.download(id)} fetches the raw file from one of four source kinds, applies the six-step recipe, checks the result against \code{expected.json} --- a mismatch raises \code{ContractError} and nothing is cached --- and caches the table as Parquet with its provenance record. Bottom: a model is a single \code{fit\_fn} (a \code{predict\_fn} is accepted and wrapped), called by both protocols; every call and failure is counted, and \code{fleet\_summary} returns \code{auc\_mean} only when every dataset was scored, every table met its contract and the scores were row-independent.}
\label{fig:architecture}
\end{figure}

\pamir{} ships the catalogue, the recipes and the evaluation code, never the data. \code{pamir.download(id)} fetches the raw file from the location recorded in the catalogue --- a Kaggle dataset, a UCI archive, a GitHub file or archive, or a Hugging Face dataset repository --- and applies the recipe locally. Every source is pinned: Kaggle datasets to a version, the Hugging Face repository to a revision, GitHub files to a commit, and the UCI archives, which carry no version, by the digest of the file. Twelve datasets are hosted on Kaggle; \code{kagglehub} 1.0.2 fetched all twelve without an account in our tests, while older versions need Kaggle credentials. The other seven come from UCI, GitHub and Hugging Face, and \code{pamir.download\_open()} fetches exactly those. Each rebuilt table is checked against a frozen contract, \code{expected.json}, that records its feature names and types, its row count, its number of defaults, the SHA-256 digest of the raw file and the digest of the target vector in stored order. A mismatch raises an error and nothing is cached, so a download that does not reproduce the contract fails rather than silently producing a different benchmark (\code{strict=False} turns the error into a warning, for development). The check is repeated whenever a cached table is loaded, and the provenance record --- source, snapshot and digests --- is stored in the Parquet metadata. All 19 tables were rebuilt from their pinned snapshots and met their contracts on 2 October 2026. Appendix~\ref{app:provenance} lists, for every dataset, the source fetched, its snapshot and the target definition, and Figure~\ref{fig:architecture} traces the pipeline from catalogue entry to fleet summary.

\subsection{Harmonization recipe}
\label{sec:harmonization}

\begin{figure}[tp]
\centering
\includegraphics[width=\linewidth]{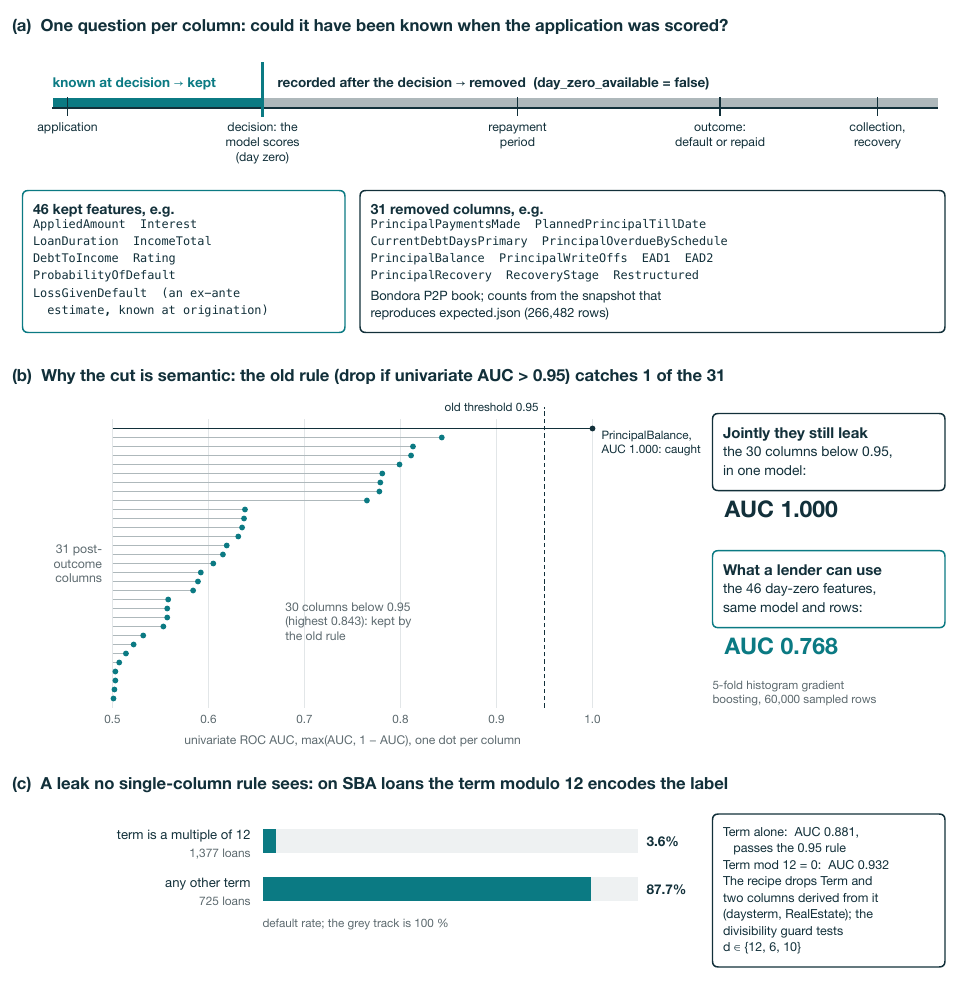}
\caption{\textbf{The day-zero cut asks whether a column could have been known when the application was scored; a univariate threshold misses most post-outcome columns.} (a)~Kept and removed columns of the Bondora loan book on a loan's timeline. (b)~Univariate ROC AUC of the 31 Bondora columns the cut removes: the earlier rule (drop if AUC $> 0.95$) catches one of them, yet the other 30 together predict the outcome perfectly (AUC 1.000, five-fold histogram gradient boosting on 60{,}000 rows), against 0.768 for the 46 day-zero features. (c)~On the SBA loans \citep{li2018sba} the term alone has AUC 0.881, but whether it is a multiple of 12 separates a 3.6\% from an 87.7\% default rate (indicator AUC 0.932); the recipe drops the term and the two columns derived from it. Measured on the pinned raw snapshots that reproduce \code{expected.json}.}
\label{fig:dayzero}
\end{figure}

Every dataset is rebuilt by the same six-step recipe, parameterized per dataset in the catalogue:
\begin{enumerate}
  \item read the raw file with its dataset-specific separator, encoding and, for headerless or ARFF sources, declared column names;
  \item derive the binary target \code{\_\_target\_\_} (1 = default) by a numeric parse or a dataset-specific rule;
  \item drop the target, identifier, date and unnamed-index columns;
  \item coerce numeric-looking strings, drop constant and all-missing columns, and drop surrogate-key columns, detected by content (one distinct value per row) rather than by name;
  \item drop every column that the recipe marks as not available to the lender at the moment of decision (\code{day\_zero\_available = false}), plus dataset-specific drops;
  \item shuffle the rows with a fixed seed.
\end{enumerate}
Step~5, the \emph{semantic day-zero cut}, replaced an earlier statistical rule that dropped any column with univariate AUC above 0.95. The statistical rule passed sets of individually weak columns that leak the outcome jointly; the semantic rule asks instead whether a column could have been known when the application was scored (Figure~\ref{fig:dayzero}). On the Bondora loan book, for example, the cut removes 31 post-outcome columns and keeps Bondora's own origination estimates of default probability and loss given default; on the SBA loan data \citep{li2018sba} the recipe removes the loan term, because the term modulo 12 encodes the label, together with two columns derived from it. Step~6 exists because file order in the sources is an export artefact, not an origination sequence (Section~\ref{sec:limitations}). A few datasets need a dataset-specific repair: the \code{lc\_my} credit score is inflated tenfold on a subset of rows and is rescaled, and the headerless PAKDD file is rebuilt from the competition's variable list \citep{pakdd2010}.

\subsection{Leakage audit}
\label{sec:audit}

\begin{figure}[tp]
\centering
\includegraphics[width=\linewidth]{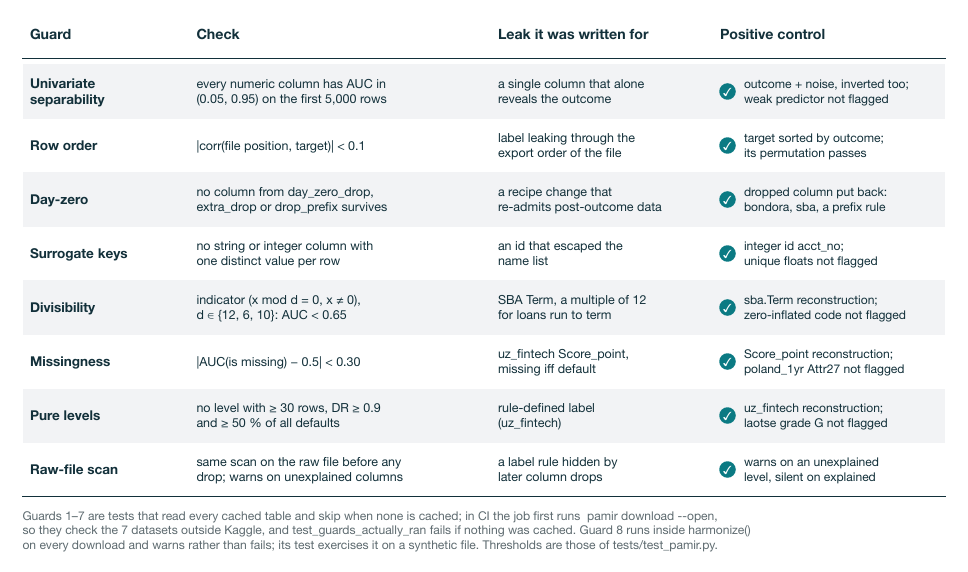}
\caption{\textbf{The eight leakage guards, their thresholds and their positive controls.} Each row gives the check, the leak it was written for, and the test that reconstructs that leak and asserts that the guard fires, together with the case that must not fire. Guards 1--7 read every cached table and run in continuous integration on the seven datasets outside Kaggle; guard 8 runs inside the harmonizer on every download and warns rather than fails.}
\label{fig:guards}
\end{figure}

The audit that produced the collection is encoded as tests that run in continuous integration on every change:
\begin{itemize}
  \item \textbf{univariate separability}: no numeric column has an AUC outside $(0.05, 0.95)$ against the target;
  \item \textbf{row order}: the correlation between file position and target is below 0.1 in absolute value;
  \item \textbf{day-zero}: no column on a dataset's day-zero drop list survives harmonization;
  \item \textbf{surrogate keys}: no column is a per-row identifier;
  \item \textbf{divisibility}: no integer column encodes the label through a modular pattern, the SBA failure mode;
  \item \textbf{missingness}: no missingness indicator predicts the target beyond what is documented as a legitimate property of the data;
  \item \textbf{pure levels}: no categorical level is almost entirely defaults, the signature of a rule-defined label;
  \item \textbf{raw-file scan}: before any column is dropped, the raw source is scanned for rule-defined labels, and unexplained hits are reported.
\end{itemize}
A guard that never fires is indistinguishable from a broken guard, so each one has a \emph{positive control}: a test that reconstructs the leak the guard was written for and asserts that the guard catches it, paired with a negative control that must not fire (Figure~\ref{fig:guards}). The data-dependent guards run in continuous integration on the seven datasets outside Kaggle and on all 19 wherever the data are cached.

\begin{figure}[tp]
\centering
\includegraphics[width=\linewidth]{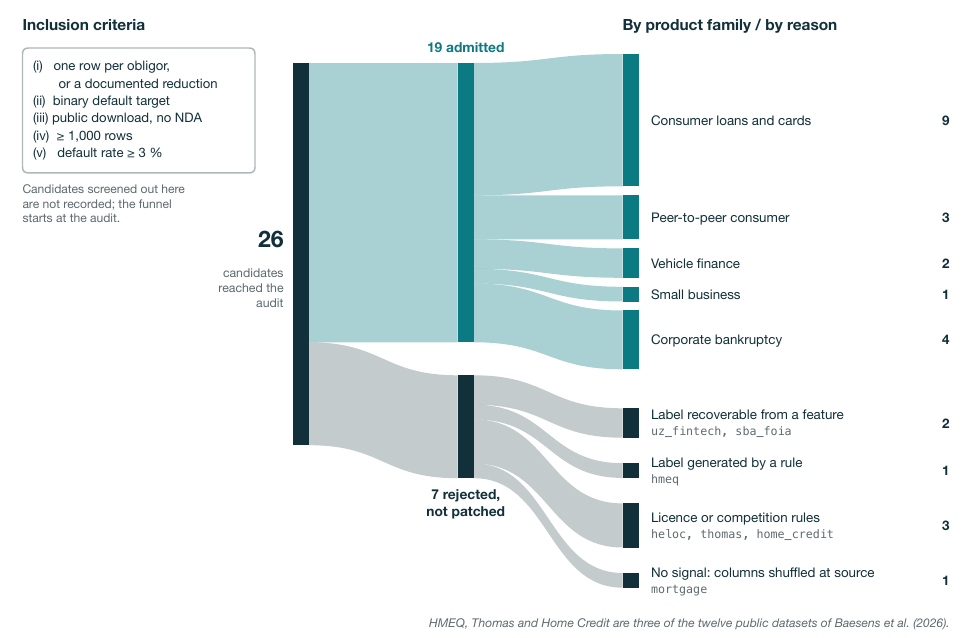}
\caption{\textbf{The audit admitted 19 datasets and rejected seven candidates rather than patching them.} Band widths are proportional to the number of datasets. Admitted datasets are shown by product family (as in Table~\ref{tab:datasets}), rejected ones by the reason for rejection. Candidates screened out earlier by the inclusion criteria of Section~\ref{sec:collection} are not recorded. Three of the rejected sets --- HMEQ, Thomas and Home Credit --- are among the twelve public datasets of \citet{baesens2026foundation}.}
\label{fig:funnel}
\end{figure}

The audit rejected seven candidates rather than patching them: two whose label is recoverable from a feature (\code{uz\_fintech}, \code{sba\_foia}), one whose label is generated by a rule (\code{hmeq}), three whose licence or competition rules forbid the use \pamir{} makes of them (\code{heloc}, \code{thomas}, \code{home\_credit}), and one whose columns were shuffled independently at the source, so that no signal survives (\code{mortgage}). Figure~\ref{fig:funnel} traces the audited candidates to the collection. Some of the excluded sets are in use elsewhere: HMEQ, Thomas and Home Credit are among the twelve public datasets of \citet{baesens2026foundation}, and HMEQ, HELOC and Home Credit among the credit tasks of BeyondArena. BeyondArena has since retired HMEQ: its curation record reports near-duplicate clusters and number-formatting and missingness artefacts that separate the two classes on their own (AUC 0.953) \citep{datafoundry2026}, consistent with the rule-generated label that led \pamir{} to exclude it.

\subsection{Licensing and redistribution}
\label{sec:licensing}

The \pamir{} code is licensed under Apache-2.0; that licence covers the code only. Because \pamir{} never redistributes a table, each user downloads each dataset from its source and is bound by that source's terms. The catalogue records, per dataset, the licence declared by the downloaded source and the obligation it creates. Two caveats apply. First, a licence shown on a Kaggle re-upload is set by the re-uploader and does not transfer rights in the underlying data; Table~\ref{tab:datasets} marks these cases. Second, attribution is required by licence for six datasets whose primary source is a CC-BY-4.0 UCI record (\code{bankruptcy}, \code{poland\_1yr}, \code{poland\_3yr}, \code{poland\_5yr}, \code{south\_german}, \code{taiwan}) and for \code{lc\_small} (ODbL, share-alike). \code{pamir.dataset\_info(id)} returns the citation for each dataset; the canonical references are \citet{liang2016}, \citet{zieba2016}, \citet{gromping2019}, \citet{yeh2009}, \citet{li2018sba}, \citet{kaggle2011gmsc} and \citet{pakdd2010}. Results computed on any dataset may be published.

\subsection{A living collection}
\label{sec:living}

\begin{figure}[tp]
\centering
\includegraphics[width=\linewidth]{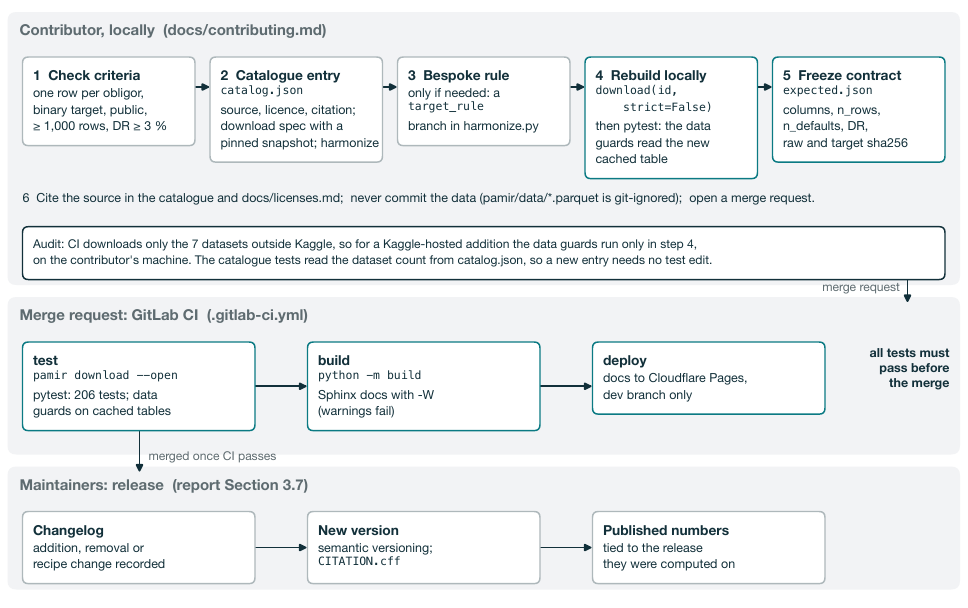}
\caption{\textbf{A dataset enters \pamir{} as a recipe, never as a file.} A contributor adds a catalogue entry with a pinned snapshot, rebuilds the table locally, runs the test suite on it and freezes the result as the \code{expected.json} contract; continuous integration reruns the tests, builds the package and documentation, and the merged change is recorded in the changelog and shipped as a new version. Continuous integration downloads only the seven datasets outside Kaggle, so the data guards on a new Kaggle-hosted dataset run on the contributor's machine.}
\label{fig:contributing}
\end{figure}

\pamir{} is maintained as a living collection rather than a one-off release. A dataset is added as a recipe, never as a file: a contributor adds a catalogue entry with its source, licence, citation, a download specification pinned to a snapshot and harmonization rules, rebuilds the table locally, and freezes the validated result --- including the digests of the raw file and of the target --- in \code{expected.json}, which becomes the contract that every later download is checked against. The leakage guards of Section~\ref{sec:audit} and the rest of the test suite must pass on the new table before the entry is merged; continuous integration reruns them on the seven datasets outside Kaggle, so for a Kaggle-hosted addition the contributor runs them locally and attaches the output to the merge request (Figure~\ref{fig:contributing}). Additions, removals and recipe changes are recorded in the changelog and shipped as new versions, so that a published number can be tied to the release it was computed on. The public probability-of-default datasets used in recent studies but not yet in \pamir{} \citep{baesens2026foundation} are the first candidates for the next releases; each will enter only if it passes the same audit.

\section{Evaluation protocols}
\label{sec:protocols}

\subsection{Model interface and metric}

A model is a single function. The streaming protocol calls it as \code{fit\_fn(X\_train, y\_train)}, which returns a \code{scorer(X\_rows)} that gives one score per row, higher meaning more likely to default. The i.i.d.\ protocol calls the equivalent \code{predict\_fn(X\_train, y\_train, X\_test)}; a \code{predict\_fn} passed to the streaming protocol is wrapped (\code{from\_predict\_fn}) and called once per scoring batch. Features arrive with their raw types --- most datasets carry object or Boolean columns --- and nothing is imputed or encoded by the harness. The baselines encode non-numeric columns against the levels seen in the training rows (\code{pamir.baselines.fit\_encoder}), giving an unseen level its own code, so that a row's score does not depend on the other rows being scored. The shared-level encoder of release 0.3.0, \code{encode\_features}, remains legal in the i.i.d.\ protocol, which hands the model the rows it must score, but not in the streaming protocol, where it would let a row's code depend on rows that arrived after it. Computing target statistics on the scoring rows is never legal. The metric is ROC AUC (the Gini coefficient, $2\,\mathrm{AUC}-1$, is reported alongside it), and the fleet metric is the unweighted mean over the 19 datasets.

\subsection{i.i.d.\ protocol}

The i.i.d.\ protocol reproduces the evaluation of general tabular benchmarks, so that \pamir{} numbers can be set against theirs. Each dataset is permuted with a seed, the first 70\% of rows train the model and the rest are scored once; the AUC is averaged over five seeds and reported with its standard deviation. By default the i.i.d.\ protocol uses the full table.

\subsection{Streaming protocol}
\label{sec:streaming}

\begin{figure}[tp]
\centering
\includegraphics[width=\linewidth]{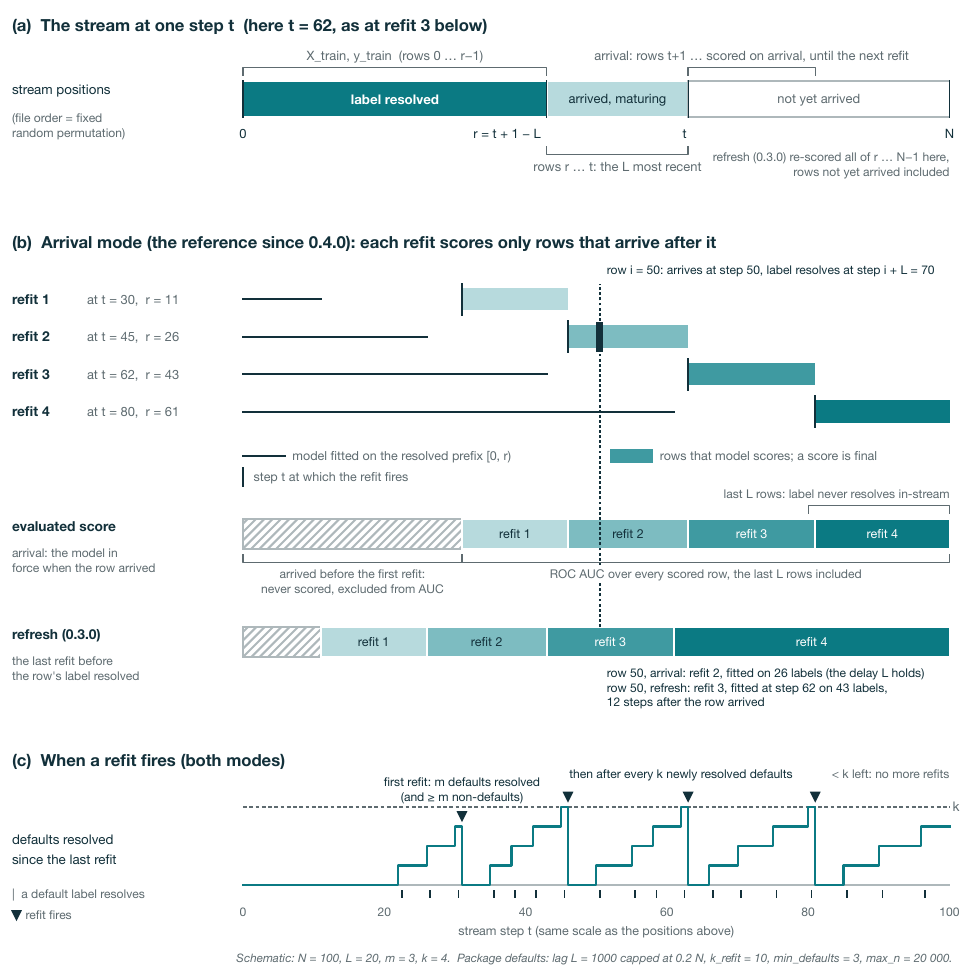}
\caption{\textbf{In arrival mode a refit's model scores only the applications that arrive after it, so the label delay binds for every row.} Schematic stream with $N = 100$ rows, delay $L = 20$, warm-up $m = 3$ and cadence $k = 4$ (package defaults: $L = 1000$ capped at $0.2\,N$, $k = 10$, $m = 3$, \code{max\_n} $= 20{,}000$). (a)~At step $t$ the labels of rows $0, \dots, r-1$, with $r = t + 1 - L$, are resolved and form the training set; the $L$ most recent rows have arrived but not matured. (b)~Each refit is fitted on its resolved prefix (line) and scores the rows that arrive after it, up to the next refit (bar); a score, once given, is final. The two strips show which refit owns each row's evaluated score in arrival mode and, for comparison, in the refresh mode of release 0.3.0, where every refit re-scored all rows from $r$ onwards, including rows that had not yet arrived. Row~50 is scored in arrival mode by refit~2, fitted on 26 labels; in refresh mode by refit~3, fitted on 43 labels twelve steps after the row arrived. Rows that arrive before the first refit are never scored and are excluded from the ROC AUC; the last $L$ rows are scored, but their labels never resolve inside the stream. (c)~Both modes share the trigger: the first refit fires once $m$ defaults (and at least $m$ non-defaults) have resolved, every later one after $k$ newly resolved defaults, and none at the final step.}
\label{fig:streaming}
\end{figure}

The streaming protocol replays a dataset as a stream of applications in stored order (Algorithm~\ref{alg:stream}, Figure~\ref{fig:streaming}). At step $t$ application $t$ arrives, and the labels of rows before position $r = \max(0, t+1-L)$ have resolved; the labels of the $L$ most recent rows are still maturing. A refit is triggered once $m$ defaults have resolved, and thereafter whenever $k$ new defaults have resolved since the previous refit, provided that at least $m$ non-defaults have also resolved. At a refit the model is fitted on the resolved prefix, and the scorer it returns scores every application that arrives after that step and before the next refit. A score, once given, is final. The model never sees the features of an application that has not arrived, nor a label that had not matured when the application it scores arrived, so the delay $L$ binds for every row. Applications that arrive before the first refit are not scored. The reported value is the ROC AUC over all scored rows, including the last $L$, whose labels resolve only after the stream ends.

Refitting on resolved defaults, rather than on a fixed number of rows or on wall-clock time, ties the update schedule to the arrival of information: a book with a 3\% default rate is refitted far less often, per row, than one with a 40\% rate.

For speed, the scorer is called on batches of applications that have already arrived, and it must score each row independently of the others in its batch. The evaluator re-scores a random subset of up to 64 rows of every batch on its own; if any score changes beyond a numerical tolerance, the run is flagged \code{row\_independent = False} and receives no fleet mean (\code{batch\_scoring=False} scores one row at a time). If a refit fails, the previous scorer stays in force, and the failure is counted.

\paragraph{Label budget.} Each score is recorded with the number of resolved labels behind it, $r$ at the refit that produced its scorer. Besides the overall AUC, the evaluator reports the AUC of the rows scored with fewer than 100, 100--299, 300--999, 1{,}000--2{,}999, 3{,}000--9{,}999 and at least 10{,}000 labels (a group with too few rows or defaults has no value), and the cumulative AUC over resolved rows at every refit, which gives a learning curve per dataset. The label-budget AUCs answer the question the protocol is built for: how well a model ranks applicants when only a few hundred outcomes have matured.

\paragraph{Delay cap.} A delay as long as the stream leaves nothing to learn from: at $L = 1000$ the two 1{,}000-row datasets would never be scored. The effective delay is $\min(L, \lfloor 0.2\,n \rfloor)$ for a stream of $n$ rows (\code{max\_lag\_frac=0.2}) and is reported per dataset as \code{lag\_effective}. At the reference setting it is shorter than $L$ for the four datasets with fewer than 5{,}000 rows: 200 for \code{conorsully} and \code{south\_german}, 420 for \code{sba} and 890 for \code{gastonstat}.

\paragraph{Refresh mode.} Release 0.3.0 used a different rule, kept as \code{mode="refresh"}: at every refit the model re-scored every row whose label was unresolved, including rows that had not yet arrived, whose features it therefore saw in advance, and a row was evaluated on the last score committed before its label resolved. Under that rule the delay binds mainly for the last $L$ rows of each stream, and a row's final score typically comes from a model fitted on almost every row that precedes it (Figure~\ref{fig:streaming}). Numbers of release 0.3.0 are reproduced with \code{mode="refresh"}, \code{max\_lag\_frac=None} and the \code{*\_v03} baselines; refresh-mode and arrival-mode numbers are not comparable.

\begin{algorithm}[t]
\caption{\textbf{Streaming evaluation in \pamir{} 0.4.0, arrival mode} (\code{pamir.evaluate\_one}).}
\label{alg:stream}
\begin{algorithmic}[1]
\Require rows $(x_i, y_i)$, $i = 0, \dots, N-1$, in stored order; model \code{fit\_fn} $g$; delay $L$; refit cadence $k$; warm-up $m$
\State $L \gets \min(L, \lfloor 0.2\,N \rfloor)$; \quad $\hat{s}_i \gets \text{undefined}$ for all $i$; \quad $h \gets \text{none}$; \quad $D \gets 0$; \quad $D_{\text{new}} \gets 0$
\For{$t = 0, 1, \dots, N-1$}
  \If{$h \ne \text{none}$}
    \State $\hat{s}_t \gets h(x_t)$ \Comment{row $t$ is scored on arrival; the score is final}
  \EndIf
  \State $r \gets \max(0,\, t+1-L)$ \Comment{rows $0,\dots,r-1$ have resolved labels}
  \If{$t \ge L$ \textbf{and} $y_{t-L} = 1$}
    \State $D \gets D + 1$; \quad $D_{\text{new}} \gets D_{\text{new}} + 1$
  \EndIf
  \If{a refit is due \textbf{and} $r > 0$ \textbf{and} $t + 1 < N$}
    \State $h \gets g\big(x_{0:r},\, y_{0:r}\big)$ \Comment{scores rows $t+1, t+2, \dots$ until the next refit}
    \State $D_{\text{new}} \gets 0$
  \EndIf
\EndFor
\State \Return $\mathrm{AUC}\big(\{(y_i, \hat{s}_i) : \hat{s}_i \text{ defined}\}\big)$
\end{algorithmic}
\end{algorithm}

\begin{table}[t]
\centering
\caption{\textbf{Streaming-protocol parameters.} The defaults form the reference setting.}
\label{tab:params}
\small
\begin{tabular}{@{}llp{0.62\linewidth}@{}}
\toprule
Parameter & Default & Meaning \\
\midrule
\code{mode} & \code{arrival} & \code{arrival}: each row is scored when it arrives; \code{refresh}: the rule of release 0.3.0 \\
\code{lag} ($L$) & 1000 & label-maturation delay, in stream positions \\
\code{max\_lag\_frac} & 0.2 & cap on the delay: $L_{\text{eff}} = \min(L, \lfloor 0.2\,n \rfloor)$; \code{None} disables it \\
\code{k\_refit} ($k$) & 10 & refit after every $k$ newly resolved defaults \\
\code{max\_n} & 20000 & truncate each stream to its first \code{max\_n} rows \\
\code{min\_defaults} ($m$) & 3 & resolved defaults (and non-defaults) required before the first refit \\
\code{on\_error} & \code{warn} & policy when the model raises: \code{warn}, \code{raise} or \code{ignore}; failures are counted under all three \\
\bottomrule
\end{tabular}
\end{table}

\paragraph{What the streaming score measures.}
Because rows are stored in a fixed random permutation, the stream is exchangeable: there is no calendar drift, and the protocol measures how well a model ranks applicants from a cold start on a growing, delayed and schedule-limited label set --- not robustness to temporal shift. In arrival mode the delay binds for every row: a row scored at position $i$ by a refit at step $t < i$ is ranked by a model fitted on $\max(0, t+1-L)$ labels. The label-budget AUCs separate the cold-start regime, in which few outcomes have matured, from the regime in which the training set is large; the overall AUC mixes the two in proportions set by the stream length, the delay and the default rate.

\subsection{Reference setting and reporting}

The streaming parameters change the score, not only the runtime. More frequent refits give fresher models: in the package walkthrough setting, a logistic regression on \code{taiwan} (\code{lag}=500, \code{max\_n}=5000) scores 0.7213, 0.7174 and 0.7132 in arrival mode at $k = 10$, 40 and 100, with 105, 27 and 11 refits respectively. Two runs are comparable only when \code{mode}, \code{lag}, \code{max\_lag\_frac}, \code{k\_refit} and \code{max\_n} all match. The package defaults (Table~\ref{tab:params}) are the reference setting, and any reported number should state them and any deviation. At \code{max\_n}=20{,}000, ten of the 19 datasets are truncated to the first 20{,}000 rows of their permuted table. Because the i.i.d.\ protocol uses the full table by default, a comparison between the two protocols should fix \code{max\_n} to the same value in both.

\subsection{Coverage is part of the result}
\label{sec:coverage}

If the model raises, the affected rows go unscored and the dataset may yield no AUC. Averaging over the datasets that remain reports a number the model has selected by failing: a model that crashes on the hardest datasets can read \emph{higher} than one that scores them all. \pamir{} therefore counts every call and every failure, and \code{pamir.fleet\_summary} reports the headline mean only for a complete run on tables that met their contract, with scores that passed the row-independence probe; the partial mean is exposed separately as a diagnostic that is not comparable across models. In the setting of the package walkthrough, six datasets with \code{lag}=500, \code{k\_refit}=40 and \code{max\_n}=5000, a logistic regression that scores all six has a mean AUC of 0.7177 in arrival mode, while a variant that fails on the four datasets with more than 15 features would report 0.7294 over the two it survives. A submission must report full coverage, or it is not a fleet result.

\section{Leakage-controlled synthetic augmentation}
\label{sec:synthetic}

Synthetic data are often proposed as a remedy for small, imbalanced credit books, and the claimed benefit is rarely measured under conditions that would reveal an artefact. The module \code{pamir.synthetic} answers one question under a fixed protocol: does adding synthetic rows to the training data improve discrimination on held-out real rows, and how faithful are those rows?

\subsection{Estimand and composition}

\begin{figure}[tp]
\centering
\includegraphics[width=\linewidth]{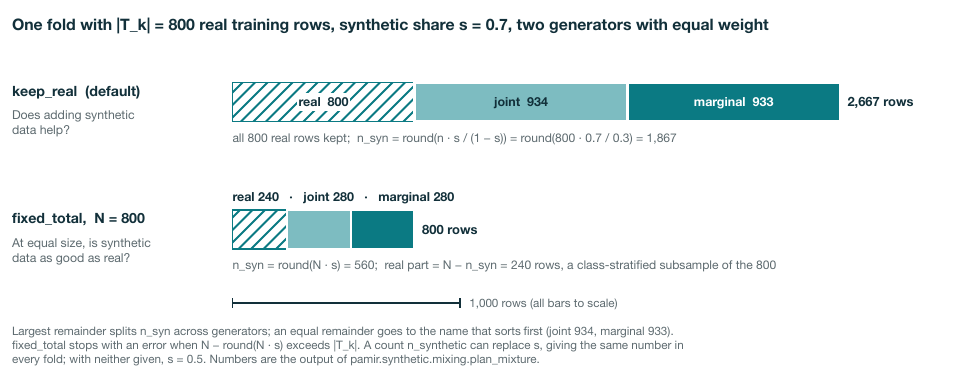}
\caption{\textbf{The two sizings of a synthetic mix answer different questions.} One fold with 800 real training rows, synthetic share $s = 0.7$ and two equally weighted generators, drawn to scale. Under the default sizing all real rows are kept and $n_{\text{syn}} = \mathrm{round}\big(n\,s/(1-s)\big) = 1{,}867$ synthetic rows are added; under fixed-total sizing the frame is pinned to $N = 800$ rows, of which 560 are synthetic and 240 are a class-stratified subsample of the real rows. The synthetic part is split by the largest-remainder method, with ties broken by generator name.}
\label{fig:sizing}
\end{figure}

Let $\mathcal{D}$ be a labelled table, $\mathcal{G}$ a generator and $\pi$ an evaluation protocol. The quantity of interest is
\begin{equation}
\Delta(\mathcal{G}, \pi) = \mathrm{AUC}_\pi(\mathcal{D} \cup \mathcal{S}) - \mathrm{AUC}_\pi(\mathcal{D}), \qquad \mathcal{S} \sim \mathcal{G},
\end{equation}
which is interpretable only if $\mathcal{G}$ was fitted on rows disjoint from those on which $\mathrm{AUC}_\pi$ is computed. Two parameters govern the mix: the synthetic share $s \in [0,1)$ (0.5 by default) and generator weights $w_g$, normalized to $\tilde{w}_g$. Under the default sizing every real training row is kept and $n_{\text{syn}} = \mathrm{round}\big(n\,s/(1-s)\big)$ synthetic rows are added, which answers ``does adding synthetic data help?''. Under the alternative, fixed-total sizing, the frame is pinned to $N$ rows and the real part is subsampled, which answers ``at equal size, is synthetic data as good as real data?'' (Figure~\ref{fig:sizing}). The synthetic part is split across generators by the largest-remainder method, so the per-generator counts always sum to $n_{\text{syn}}$, with ties broken deterministically by name. Real subsamples are stratified by class, and a target rate can be imposed on the synthetic part when the aim is to rebalance a low-default book.

\subsection{The leakage contract}

\begin{figure}[tp]
\centering
\includegraphics[width=\linewidth]{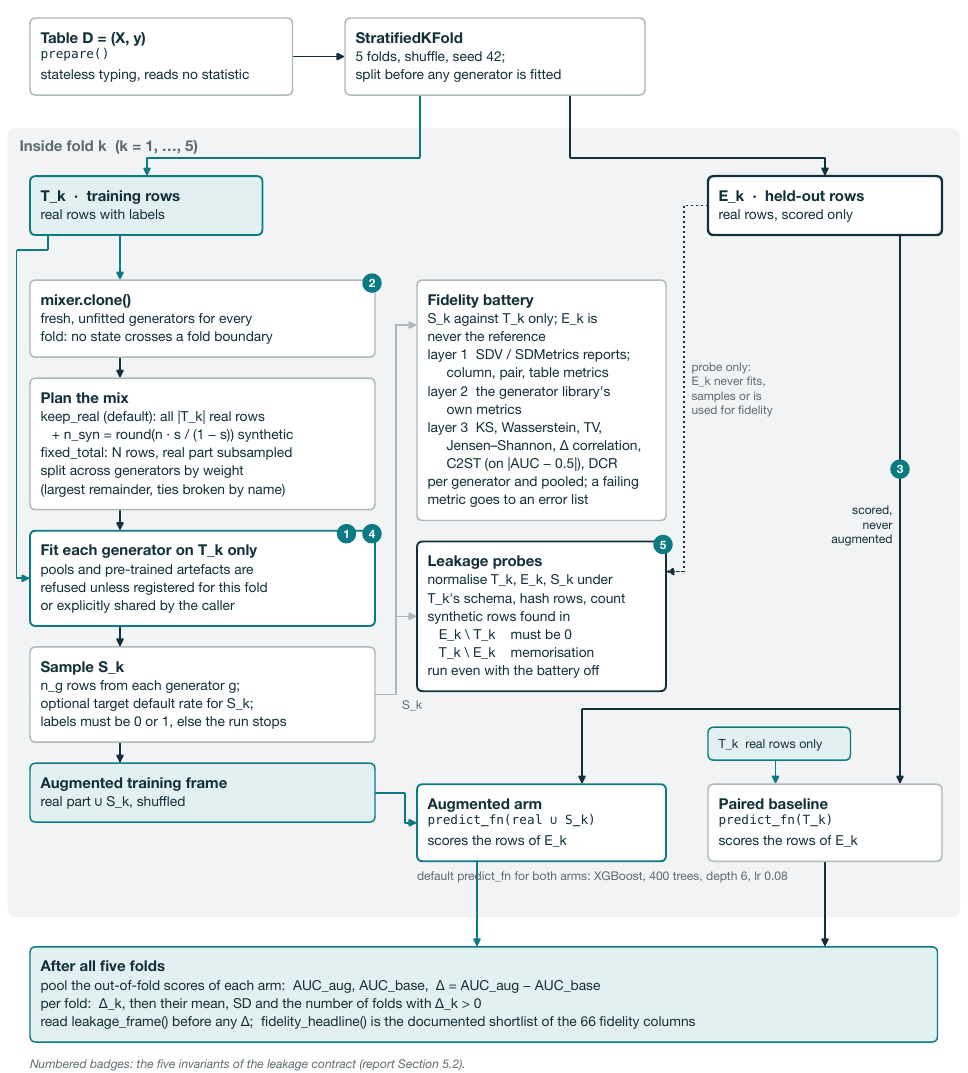}
\caption{\textbf{Inside each cross-validation fold, generators see only the training rows, and the held-out rows are scored and probed but never augmented.} The table is typed statelessly and split by stratified five-fold cross-validation before any generator is fitted. In fold $k$, fresh generators are fitted on the training rows $T_k$ alone, the planned synthetic rows $S_k$ are sampled, and the augmented frame trains the augmented arm while the paired baseline trains on $T_k$ alone; both arms score the held-out rows $E_k$. The fidelity battery compares $S_k$ with $T_k$ only. $E_k$ enters the leakage probes, and nothing else, as a comparison set: rows are hashed under $T_k$'s schema, and a synthetic row that matches a held-out-only row invalidates the result. After five folds the out-of-fold scores of each arm are pooled, and $\Delta$ is reported both pooled and per fold. Numbered badges mark the five invariants of the leakage contract.}
\label{fig:synthfold}
\end{figure}

Five invariants hold for every fold $k$ with training rows $T_k$ and held-out rows $E_k$ (Figure~\ref{fig:synthfold}):
\begin{enumerate}
  \item a generator observes only $\mathcal{D}[T_k]$: the fold is split before any generator is fitted;
  \item no state crosses a fold boundary: generators are cloned and reset per fold, so information from fold $k{+}1$'s training rows, most of which are fold $k$'s test rows, cannot flow back;
  \item $\mathcal{D}[E_k]$ is never augmented: synthetic rows enter the training argument only;
  \item pre-computed material is fold-bound: a pool of pre-generated rows or a pre-trained generator artefact is refused unless it is registered per fold or the caller explicitly vouches for sharing it;
  \item violations are detectable after the fact: every fold reports how many synthetic rows exactly reproduce a held-out row that is not also a training row.
\end{enumerate}
The probe in the last invariant hashes rows after normalizing all three frames under the real table's schema. Without that step, a generator that returns an integer column as floats, or a pool read back from CSV as text, would reproduce held-out rows verbatim and still read as clean. Counts are taken over synthetic rows rather than distinct hashes, so a generator that emits one memorized row a hundred times is charged a hundred times. The probes run even when the fidelity battery is switched off, because they cost milliseconds and a delta reported without them cannot be checked.

\subsection{Fidelity battery and protocol}

The fidelity report has three layers, and a metric that fails lands in an error list without aborting the report. The first layer runs the SDV/SDMetrics reports and per-column, per-pair and per-table metrics, including detection metrics and machine-learning-efficacy arms \citep{patki2016}. The second collects any metrics the generator's own library provides. The third needs only NumPy, pandas, SciPy and scikit-learn \citep{pedregosa2011sklearn}: per-column Wasserstein, total-variation and Jensen--Shannon distances, the change in the correlation matrix, a classifier two-sample test \citep{lopezpaz2017c2st} and distances to the closest real record. The classifier two-sample AUC is read on a two-sided scale: 0.5 means indistinguishable, and values below 0.5 indicate duplication of real rows rather than better-than-real quality, so generators are ranked on $|\mathrm{AUC} - 0.5|$. A full run yields 66 fidelity columns per fold and generator; a documented shortlist of seven is intended to be read first.

The default protocol is five-fold stratified cross-validation with an XGBoost classifier \citep{chen2016xgb} (400 trees, depth 6, learning rate 0.08), out-of-fold predictions pooled into one AUC, and a \emph{paired baseline}: the same folds are run on the real training rows alone. Without \code{xgboost} installed the harness falls back to scikit-learn's histogram gradient boosting, so the instrument check below requires \code{xgboost}. $\Delta$ is reported both as a difference of pooled out-of-fold AUCs and as a fold-wise mean with its standard deviation and the number of positive folds. Any \code{predict\_fn} with the benchmark signature can replace the classifier. Adapters wrap any SDV synthesizer, a pool of pre-generated rows, or any object with \code{fit} and \code{sample} methods; the package also contains adapters for zGAN and zEDGE, two generators developed at zypl.ai, which are distributed separately and are not evaluated in this report (see the statement of conflict of interest).

\subsection{An instrument check}

The harness was validated end to end on \code{south\_german} (1,000 rows, 20 features, default rate 0.300) with two surrogate generators chosen to have known, opposite defects: \emph{joint}, which resamples training rows and adds Gaussian jitter at 5\% of each column's standard deviation, preserving dependence but violating the schema of every integer-coded categorical; and \emph{marginal}, which resamples each column independently, preserving every marginal and destroying the dependence structure. The mix used $s = 0.7$ with equal weights, five stratified folds and the XGBoost arm. The combined mix lowered the pooled out-of-fold AUC from 0.7669 to 0.7438 ($\Delta = -0.0231$; fold-wise mean $-0.0273$, standard deviation 0.0619, two of five folds positive), so the fold spread is more than twice the effect --- which is why the paired, per-fold report is the default. Table~\ref{tab:instrument} shows the fidelity layer on the same run. SDV's aggregate quality score ranks \emph{marginal} above \emph{joint} (0.953 against 0.848), although \emph{marginal} is the generator that destroys the dependence structure (Pearson correlation similarity 0.888 against 0.988): an aggregate built mostly from marginals cannot express a copula failure. The two-sample test separates \emph{joint} from real rows perfectly (AUC 1.000), because the jitter leaves a fingerprint no marginal statistic reveals, and the closest-record distance separates the two failure modes --- \emph{joint} crowds the real rows, \emph{marginal} lies far from them. The leakage probes were clean in all ten generator--fold pairs.

\begin{table}[t]
\centering
\caption{\textbf{The fidelity battery exposes the defects of two surrogate generators that the aggregate SDV score misranks.} Fidelity on \code{south\_german}, averaged over five folds. \emph{joint}: row resampling with jitter (dependence kept, schema violated); \emph{marginal}: independent column resampling (marginals kept, dependence destroyed).}
\label{tab:instrument}
\small
\begin{tabular}{@{}lrr@{}}
\toprule
Metric & \emph{joint} & \emph{marginal} \\
\midrule
SDV quality score & 0.8484 & 0.9527 \\
SDV diagnostic score & 0.8797 & 1.0000 \\
KS complement (column mean) & 0.9388 & 0.9742 \\
TV complement (column mean) & 0.0553 & 0.9821 \\
Pearson correlation similarity (pair mean) & 0.9877 & 0.8882 \\
Classifier two-sample AUC & 1.0000 & 0.8527 \\
Mean distance to closest real record & 0.2219 & 3.4168 \\
Synthetic rows matching a held-out-only row & 0 & 0 \\
\bottomrule
\end{tabular}
\end{table}

The collapse of the total-variation complement for \emph{joint} (0.055) is a schema effect rather than a distributional one: jitter turned 17 integer-coded categoricals into continuous columns, and the metric compared a four-level distribution against one with hundreds of levels. The battery reports such type violations explicitly so that the number is not read as evidence about shape.

\section{Software and reproducibility}
\label{sec:software}

\pamir{} is a pure-Python package (\code{pamir-credit}, Python $\geq$ 3.9) whose core depends on NumPy, pandas, scikit-learn, SciPy and PyArrow and needs no GPU. Optional extras install the fetch dependencies (\code{[data]}) and the synthetic-data stack (\code{[synthetic]}: SDV, SDMetrics, XGBoost). A command-line interface lists, inspects, downloads and locates cached datasets (\code{pamir list | info | download | cache}). The following session reproduces the reference streaming run:
\begin{verbatim}
pip install "pamir-credit[data]"

from pamir import evaluate, fleet_summary, gbdt_fit
results = evaluate(gbdt_fit)    # arrival mode, lag=1000 (cap 0.2 n), k_refit=10, max_n=20000
fleet_summary(results)          # auc_mean is None unless every dataset is validly scored
\end{verbatim}
The test suite (206 tests) covers catalogue and recipe integrity and the data contract, the harmonizer and downloader, the leakage guards and their positive controls, the invariants of both streaming modes (for example, that in arrival mode a row is scored only by a model fitted on labels at least $L$ positions old and never on a later row's features), failure accounting, the reference baselines, data quality and the synthetic module. Tests that need data skip cleanly when a dataset is not cached. The documentation is built with Sphinx and additionally exported as \code{llms.txt} and \code{llms-full.txt}, so that a language-model agent can read the full API and protocol in one request. A Hugging Face dataset card makes the benchmark discoverable through Hub search; it hosts no data. Releases follow semantic versioning with a changelog and a \code{CITATION.cff} file.

\paragraph{Cost.} In arrival mode every refit is a full fit, and the rows that arrive until the next refit are scored once. At $k = 10$, \code{bondora} truncated to 5{,}000 rows takes 161 refits; with a 60-tree histogram gradient-boosting model the run takes about 16\,s on four threads of an AMD EPYC 4564P, in either mode. A full-fleet run at the reference setting took, summed over the 19 datasets, about 5 minutes for the logistic baseline (two threads per dataset) and 16 minutes for the GBDT baseline (four threads per dataset) on the same CPU; the longest single dataset, \code{bondora} with GBDT, took 3.4 minutes. The i.i.d.\ protocol on the full tables took under a minute per baseline. On the synthetic side, one full fidelity report on a $1{,}000 \times 20$ table takes about 11\,s, and five folds with two generators about two minutes, almost all of it measurement rather than generation.

\section{Reference baselines and results}
\label{sec:baselines}

Two baselines ship with the package and run on all 19 datasets without modification, each as a \code{fit\_fn} for the streaming protocol (\code{logistic\_fit}, \code{gbdt\_fit}) and as the equivalent \code{predict\_fn} (\code{logistic\_baseline}, \code{gbdt\_baseline}). The logistic baseline is a logistic regression with median imputation and standardization; the GBDT baseline is a histogram gradient-boosting classifier (150 iterations, learning rate 0.1) from scikit-learn \citep{pedregosa2011sklearn}. Both encode non-numeric columns against the training levels and return base-rate scores when a training prefix holds a single class. They are fixed reference points rather than competitive models, and nothing is tuned. The 0.3.0 baselines, which encoded against the levels of the training and scoring rows together, are kept as \code{*\_v03} to reproduce 0.3.0 numbers.

The reference runs follow an analysis plan fixed before any reference number was computed (\path{benchmarks/reference/PLAN.md}): both baselines under the arrival-mode stream at the reference setting, under the i.i.d.\ protocol on the same 20{,}000-row truncation and on the full tables, under the refresh rule of release 0.3.0, and under arrival mode with \code{lag}=250. Every run in the plan is reported, and a run that fails is reported as failed. Per dataset a run reports the AUC and Gini, the label-budget AUCs, the numbers of refits, calls, failures and scored rows, the effective delay, the contract and row-independence flags and the wall time; per run, the fleet summary, the command, the git commit and the library versions. The comparisons fixed in advance are the headroom of GBDT over logistic regression, the gap between the i.i.d.\ and the streaming protocol, the number of datasets on which the better of the two models changes between them, the label-budget curve, the effect of the delay and the difference between the arrival and refresh rules.

Table~\ref{tab:reference} gives the per-dataset results and Table~\ref{tab:budget} the label-budget curve. All 19 datasets were scored in every run, every table met its contract, every refit succeeded and every scorer passed the row-independence probe, so every fleet mean is defined. The comparisons fixed in the plan follow. Differences are taken per dataset and summarized by their mean over the 19 datasets, the number of datasets with each sign, and a paired Wilcoxon signed-rank test, reported as a description of consistency and without a correction for multiple comparisons.

\paragraph{Headroom.} Under the i.i.d.\ split on the same 20{,}000 rows, GBDT exceeds logistic regression by 0.060 in fleet mean (0.785 against 0.725) and on 14 of 19 datasets (Wilcoxon $p = 0.004$). Under the arrival-mode stream the margin shrinks to 0.036 (0.753 against 0.717), on 11 of 19 datasets ($p = 0.20$).

\paragraph{Protocol gap.} The stream scores lower than the i.i.d.\ split on all 19 datasets for GBDT, by 0.032 in fleet mean, and on 16 of 19 for logistic regression, by 0.009. In arrival mode most rows of a stream are scored by models fitted on fewer labels than the 70\% of the table that the i.i.d.\ split trains on, and the label-budget curve shows that GBDT gains more from additional labels than logistic regression does.

\paragraph{Ranking.} The better of the two baselines changes between the i.i.d.\ split and the stream on three datasets (\code{pakdd}, \code{lc\_my}, \code{lt\_vehicle}); on all three GBDT is ahead under the i.i.d.\ split and behind under the stream.

\paragraph{Label budget.} Rows scored by models fitted on fewer than 100 labels have a fleet mean AUC of 0.606 with logistic regression and 0.600 with GBDT, and rows scored with 100--299 labels 0.684 and 0.667. From 300 labels on GBDT is ahead, by 0.039 and 0.053 in the groups between 1{,}000 and 9{,}999 labels. Both models share the refit schedule, so each group holds the same rows for both.

\paragraph{Delay.} Shortening the delay from 1{,}000 to 250 positions raises the fleet mean by 0.002 for both baselines, with the shorter delay ahead on 11 (logistic regression, $p = 0.10$) and 13 (GBDT, $p = 0.08$) of 19 datasets. Once a few thousand labels have matured, 750 more or fewer change the training set of a refit by a small fraction.

\paragraph{Refresh versus arrival.} The 0.3.0 refresh rule scores higher than arrival mode, by 0.001 in fleet mean for logistic regression (12 of 19 datasets, $p = 0.59$) and by 0.003 for GBDT (16 of 19, $p = 0.023$): it scores a row with a model fitted on more labels than were available when the row arrived (Figure~\ref{fig:streaming}). At the reference setting the difference is small, and arrival-mode numbers are the ones to report.

\paragraph{Full tables.} With the full tables instead of the 20{,}000-row truncation, the i.i.d.\ fleet mean rises from 0.725 to 0.726 for logistic regression and from 0.785 to 0.794 for GBDT.

\begin{table}[tp]
\centering
\caption{\textbf{Reference results: under the arrival-mode stream the GBDT baseline leads logistic regression by less than under the i.i.d.\ split.} ROC AUC per dataset at the reference setting (Table~\ref{tab:params}) for the two package baselines, under arrival mode, under the refresh rule of release 0.3.0 and under the i.i.d.\ protocol on the same 20{,}000-row truncation (mean of five seeds). $L_{\text{eff}}$ is the effective delay. Datasets are grouped by product family as in Table~\ref{tab:datasets}. Commit \code{2c13c1b}, scikit-learn 1.9.1; per-run records, including the label-budget AUCs and refit counts, are written by \code{run\_reference.py} in \code{benchmarks/reference}.}
\label{tab:reference}
\small
\setlength{\tabcolsep}{4pt}
\begin{tabular}{@{}lrcccccc@{}}
\toprule
& & \multicolumn{3}{c}{logistic regression} & \multicolumn{3}{c}{GBDT} \\
\cmidrule(lr){3-5}\cmidrule(l){6-8}
id & $L_{\text{eff}}$ & arrival & refresh & i.i.d. & arrival & refresh & i.i.d. \\
\midrule
\texttt{conorsully} & 200 & 0.657 & 0.645 & 0.674 & 0.613 & 0.632 & 0.643 \\
\texttt{south\_german} & 200 & 0.737 & 0.751 & 0.791 & 0.730 & 0.733 & 0.783 \\
\texttt{gastonstat} & 890 & 0.771 & 0.774 & 0.777 & 0.787 & 0.800 & 0.815 \\
\texttt{lc\_small} & 1000 & 0.658 & 0.660 & 0.665 & 0.618 & 0.620 & 0.640 \\
\texttt{taiwan} & 1000 & 0.722 & 0.723 & 0.735 & 0.754 & 0.755 & 0.786 \\
\texttt{laotse} & 1000 & 0.847 & 0.848 & 0.850 & 0.921 & 0.922 & 0.938 \\
\texttt{pakdd} & 1000 & 0.594 & 0.597 & 0.603 & 0.586 & 0.595 & 0.619 \\
\texttt{lc\_my} & 1000 & 0.658 & 0.656 & 0.659 & 0.656 & 0.658 & 0.683 \\
\texttt{gmsc} & 1000 & 0.697 & 0.700 & 0.703 & 0.800 & 0.803 & 0.857 \\
\midrule
\texttt{prosper} & 1000 & 0.742 & 0.744 & 0.753 & 0.742 & 0.746 & 0.766 \\
\texttt{lc\_clean} & 1000 & 0.694 & 0.690 & 0.696 & 0.659 & 0.659 & 0.686 \\
\texttt{bondora} & 1000 & 0.700 & 0.700 & 0.706 & 0.733 & 0.735 & 0.753 \\
\midrule
\texttt{dish} & 1000 & 0.708 & 0.708 & 0.725 & 0.633 & 0.646 & 0.718 \\
\texttt{lt\_vehicle} & 1000 & 0.617 & 0.622 & 0.625 & 0.605 & 0.607 & 0.628 \\
\midrule
\texttt{sba} & 420 & 0.793 & 0.797 & 0.798 & 0.826 & 0.833 & 0.840 \\
\midrule
\texttt{poland\_5yr} & 1000 & 0.795 & 0.786 & 0.791 & 0.943 & 0.927 & 0.952 \\
\texttt{bankruptcy} & 1000 & 0.849 & 0.869 & 0.878 & 0.910 & 0.904 & 0.933 \\
\texttt{poland\_1yr} & 1000 & 0.685 & 0.681 & 0.674 & 0.930 & 0.927 & 0.952 \\
\texttt{poland\_3yr} & 1000 & 0.692 & 0.683 & 0.682 & 0.859 & 0.861 & 0.929 \\
\midrule
\textbf{Fleet mean} & & 0.717 & 0.718 & 0.725 & 0.753 & 0.756 & 0.785 \\
\bottomrule
\end{tabular}
\end{table}

\begin{table}[tp]
\centering
\caption{\textbf{Below 300 labels the two baselines are close, with logistic regression ahead in fleet mean; with more labels GBDT pulls ahead.} Fleet mean ROC AUC of the rows scored by models fitted on a given number of resolved labels (arrival mode, reference setting), with the number of datasets that reach each group. Groups differ in which datasets they contain, so the curve is descriptive.}
\label{tab:budget}
\small
\begin{tabular}{@{}lrrrr@{}}
\toprule
& \multicolumn{2}{c}{logistic regression} & \multicolumn{2}{c}{GBDT} \\
\cmidrule(lr){2-3}\cmidrule(l){4-5}
labels behind the score & AUC & datasets & AUC & datasets \\
\midrule
0--99 & 0.606 & 17 & 0.600 & 17 \\
100--299 & 0.684 & 17 & 0.667 & 17 \\
300--999 & 0.702 & 19 & 0.717 & 19 \\
1,000--2,999 & 0.722 & 17 & 0.761 & 17 \\
3,000--9,999 & 0.721 & 16 & 0.774 & 16 \\
$\geq$ 10,000 & 0.704 & 10 & 0.743 & 10 \\
\bottomrule
\end{tabular}
\end{table}

\section{Limitations}
\label{sec:limitations}

\paragraph{No calendar order.} Most sources carry no origination or resolution dates, and file order in the sources is an export artefact. \pamir{} therefore stores every table in a fixed random permutation, measures delay in stream positions rather than months, and does not model population drift. The word \emph{arrival-ordered} in the name refers to this replayed order, not to an origination sequence. The delay cap of Section~\ref{sec:streaming} shortens the delay on the four smallest datasets.

\paragraph{Batch scoring.} Arrival mode fixes a score when a row arrives, but scorers are called on batches for speed. A scorer that pools information across the rows of a batch is detected only by the row-independence probe, so a dependence that changes scores by less than its numerical tolerance, or that affects only rows outside the re-scored subset, would escape it; \code{batch\_scoring=False} removes the question at the cost of one call per row.

\paragraph{Heterogeneous data quality.} The collection mixes curated competition data (\code{gmsc}, \code{pakdd}), academic datasets (\code{taiwan}, \code{south\_german}, the Polish sets) and educational or synthetic tables (\code{conorsully}, \code{lc\_small}). Some sources are of uncertain provenance: \code{dish} is an automobile-loan table despite its Kaggle identifier, and several Kaggle re-uploads do not name the original rights holder. Sources can also change or disappear: every source is pinned to a snapshot, and if a host withdraws it, the contract fails rather than substituting a different table. The unweighted fleet mean treats all 19 datasets as equally informative, which they are not.

\paragraph{Scope of the synthetic module.} The augmentation harness uses cross-validation and does not yet run under the streaming protocol. Its leakage probe detects exact reproduction only; a generator that perturbs a held-out row slightly escapes it. Pair metrics are capped for cost, and the closest-record distance is computed on at most 2,000 rows per side, which understates memorization on larger tables.

\paragraph{Two baselines.} The reference results cover two untuned package baselines. Tuned learners and tabular foundation models are not evaluated in this release; in arrival mode a foundation model would be refitted, or re-conditioned, several hundred times per dataset (773 times on \code{bondora} at the reference setting), which calls for a GPU.

\section{Roadmap}
\label{sec:roadmap}

\pamir{} is intended to be maintained as a benchmark rather than released once. The next versions will grow the collection and address the limitations above in the following order:
\begin{enumerate}
  \item \textbf{More datasets}: the public probability-of-default datasets of recent studies that \pamir{} does not yet include \citep{baesens2026foundation}, and further sources, each admitted through the audit of Section~\ref{sec:audit}, with the aim of passing 20 datasets in the next releases.
  \item \textbf{Temporal tracks} built from sources that do carry origination dates, evaluated with out-of-time windows in the spirit of \citet{rubachev2025tabred} and with a gap between training and test periods as long as the maturation period; loan-level sources with origination dates, such as the Freddie Mac data packaged by \citet{mushava2024}, are natural candidates. Such tracks need a correction for censoring: when only loans with a terminal status are kept, recent cohorts over-represent early defaults.
  \item \textbf{Drift and tail-risk protocols}: controlled shifts of graded severity, low-default tracks and tail-focused metrics. Gaussian-prior generators are known to struggle with heavy tails \citep{pandey2025heavytail}, and credit books are where rare events carry the cost.
  \item \textbf{The synthetic harness under streaming and under shift}, since that is where synthetic data are most often claimed to help.
  \item \textbf{Reference results} beyond the two baselines: deep tabular models and tabular foundation models, and generators including copulas, CTGAN, TabSyn and TabDiff \citep{xu2019ctgan, zhang2024tabsyn, shi2025tabdiff}.
  \item \textbf{Export to general benchmarks}: the \pamir{} tables and protocols as tasks in the DataFoundry format \citep{purucker2026beyondarena}, so that a credit track can run inside a general benchmark without redistributing the data.
  \item \textbf{A frozen v1.0 protocol specification}, per-dataset datasheets \citep{gebru2021}, a leaderboard with verified submissions, and a written governance note on how the maintainers treat their own submissions.
\end{enumerate}
Although \pamir{} is credit-specific, the problems it isolates --- rare positive classes, outcomes that mature with a delay, heterogeneous and partly leaky public sources, and synthetic data whose utility is easy to overstate --- recur across tabular prediction, and we expect the protocol machinery to transfer.

\section{Availability}

Code, recipes and documentation are released under Apache-2.0. Links shown in red are provisional and will be replaced at release:
\begin{itemize}
  \item source code: \provurl{https://github.com/zypl-ai/pamir-credit};
  \item package: \provisional{\code{pip install pamir-credit}} (PyPI);
  \item documentation: \provurl{https://pamir-docs.pages.dev};
  \item Hugging Face dataset card: \provisional{URL to be announced}.
\end{itemize}
\pamir{} is updated through versioned releases; this report describes release 0.4.0 and will be revised as the collection grows. Users of \pamir{} should cite this report and the original source of every dataset they use (\code{pamir.dataset\_info(id)["citation"]}).

\section*{Conflict of interest}
\pamir{} is developed at zypl.ai, with which all authors are affiliated and which also develops synthetic-data generators for credit risk, among them zGAN and zEDGE. The package contains adapters for these two generators so that they can be measured by the same public protocol as any other; the generators themselves are distributed separately, and neither this report nor the reference results evaluate them. Any future evaluation of a model or generator developed by the maintainers will use the published protocol, ship with its code, and be marked as such.

\section*{Acknowledgements}
\textcolor{red}{We are especially grateful to zypl.ai Corp.\ for the opportunity to develop the open part of this domain.}

\bibliographystyle{unsrtnat}
\bibliography{references}

\appendix
\section{Dataset provenance}
\label{app:provenance}

Table~\ref{tab:provenance} lists, for every dataset, the source that \code{pamir.download} fetches and the definition of the default label.

\begin{table}[h]
\centering
\caption{\textbf{Every table is fetched from a named source and rebuilt locally.} Provenance, pinned snapshot and target definition of each dataset, as fetched by \texttt{pamir.download} (catalog fields \texttt{download.locator}, \texttt{version} and \texttt{revision}). Every raw file is also pinned by its SHA-256 digest in \texttt{expected.json}; the UCI archives carry no version and are pinned by the digest alone. Full recipes, including the day-zero drop lists, are returned by \texttt{pamir.dataset\_info(id)}.}
\label{tab:provenance}
\scriptsize
\begin{tabularx}{\linewidth}{@{}l>{\raggedright\arraybackslash}p{0.42\linewidth}>{\raggedright\arraybackslash}X@{}}
\toprule
id & source fetched (snapshot) & target (1 = default) \\
\midrule
\texttt{conorsully} & Kaggle: \path{conorsully1/credit-score} (v1) & Default on personal loan \\
\texttt{south\_german} & UCI: dataset 573 (sha256) & Bad credit risk \\
\texttt{gastonstat} & GitHub: \path{gastonstat/CreditScoring} (commit 250c644) & Loan default \\
\texttt{lc\_small} & Kaggle: \path{itssuru/loan-data} (v1) & Not fully paid \\
\texttt{taiwan} & Kaggle: \path{uciml/default-of-credit-card-clients-dataset} (v1) & Default on next month credit card payment \\
\texttt{laotse} & Kaggle: \path{laotse/credit-risk-dataset} (v1) & Loan default \\
\texttt{pakdd} & GitHub: \path{JLZml/Credit-Scoring-Data-Sets} (commit fe7fea9) & Bad loan (PAKDD competition target) \\
\texttt{lc\_my} & Kaggle: \path{zaurbegiev/my-dataset} (v1) & Loan default \\
\texttt{gmsc} & Kaggle: \path{brycecf/give-me-some-credit-dataset} (v1) & 90+ days past due within 2 years on revolving/installment \\
\texttt{prosper} & Kaggle: \path{henryokam/prosper-loan-data} (v1) & Loan defaulted or charged off \\
\texttt{lc\_clean} & HF: \path{RPD123-byte/credit-risk-datasets} (rev.\ 6819eb4) & Loan charged off \\
\texttt{bondora} & Kaggle: \path{marcobeyer/bondora-p2p-loans} (v387) & Loan status: defaulted \\
\texttt{dish} & Kaggle: \path{saurabhbagchi/dish-network-hackathon} (v1) & Auto loan default (provenance unverified; Kaggle title 'Automobile Loan Default Dataset') \\
\texttt{lt\_vehicle} & Kaggle: \path{mamtadhaker/lt-vehicle-loan-default-prediction} (v2) & Default on the first EMI of a vehicle loan (L\&T Financial Services) \\
\texttt{sba} & Kaggle: \path{larsen0966/sba-loans-case-data-set} (v1) & SBA-backed loan charged off \\
\texttt{poland\_5yr} & UCI: dataset 365 (sha256) & Company bankruptcy within 5 year(s) (Polish firms, 2000--2013) \\
\texttt{bankruptcy} & Kaggle: \path{fedesoriano/company-bankruptcy-prediction} (v2) & Company bankruptcy per Taiwan Stock Exchange rules (Taiwan Economic Journal, 1999--2009) \\
\texttt{poland\_1yr} & UCI: dataset 365 (sha256) & Company bankruptcy within 1 year(s) (Polish firms, 2000--2013) \\
\texttt{poland\_3yr} & UCI: dataset 365 (sha256) & Company bankruptcy within 3 year(s) (Polish firms, 2000--2013) \\
\bottomrule
\end{tabularx}
\end{table}

\end{document}